\documentclass[sigconf]{acmart}

\usepackage{graphicx}
\usepackage{subcaption}
\usepackage{algpseudocode}
\usepackage{xspace}
\usepackage{tabu}
\usepackage{bm}
\usepackage{multirow}
\usepackage{enumitem}
\usepackage[ruled,vlined,linesnumbered]{algorithm2e}
\usepackage{color,soul}
\usepackage{wrapfig}
\usepackage{amsmath,multicol, mathrsfs}
\usepackage[bottom]{footmisc}
\usepackage{makecell}
\usepackage{threeparttable}
\usepackage{tabularx}
\usepackage{longtable}
\usepackage{xcolor}
\usepackage{booktabs}
\usepackage{balance}

\newcommand{\ourdata}{{SYMPCODER}}
\newcommand{\ourdatafull}{{SYMPCODER-Full}}
\newcommand{\ourdatacom}{{SYMPCODER-Common-50}}
\newcommand{\ourdatarare}{{SYMPCODER-Rare-50}}
\newcommand{\ourprompt}{{TACO}}
\newcommand{\alterprompt}{{TASI}}

\AtBeginDocument{%
  }

\copyrightyear{2025} 
\acmYear{2025} 
\setcopyright{acmlicensed}\acmConference[CHASE '25]{ACM/IEEE International Conference on Connected Health: Applications, Systems and Engineering Technologies}{June 24--26, 2025}{New York, NY, USA}
\acmBooktitle{ACM/IEEE International Conference on Connected Health: Applications, Systems and Engineering Technologies (CHASE '25), June 24--26, 2025, New York, NY, USA}
\acmDOI{10.1145/3721201.3721383}
\acmISBN{979-8-4007-1539-6/2025/06}

\begin{document}

\title{Task as Context Prompting for Accurate Medical Symptom Coding Using Large Language Models}

%
% The "author" command and its associated commands are used to define
% the authors and their affiliations.
% Of note is the shared affiliation of the first two authors, and the
% "authornote" and "authornotemark" commands
% used to denote shared contribution to the research.
\author{Chengyang He}
% \orcid{0009-0004-1119-7969}
\affiliation{%
  \institution{Stevens Institute of Technology}
  \city{Hoboken}
  \state{New Jersey}
  \country{USA}
}
\email{che14@stevens.edu}

\author{Wenlong Zhang}
\affiliation{%
  \institution{Stevens Institute of Technology}
  \city{Hoboken}
  \state{New Jersey}
  \country{USA}
}
\email{wzhang71@stevens.edu}

\author{Violet (Xinying) Chen}
\affiliation{%
  \institution{Stevens Institute of Technology}
  \city{Hoboken}
  \state{New Jersey}
  \country{USA}
}
\email{vchen3@stevens.edu}

\author{Yue Ning}
\affiliation{%
  \institution{Stevens Institute of Technology}
  \city{Hoboken}
  \state{New Jersey}
  \country{USA}
}
\email{yning5@stevens.edu}

\author{Ping Wang}
\affiliation{%
  \institution{Stevens Institute of Technology}
  \city{Hoboken}
  \state{New Jersey}
  \country{USA}
}
\email{pwang44@stevens.edu}

%
% By default, the full list of authors will be used in the page
% headers. Often, this list is too long, and will overlap
% other information printed in the page headers. This command allows
% the author to define a more concise list
% of authors' names for this purpose.
\renewcommand{\shortauthors}{He et al.}

\begin{abstract}
Accurate medical symptom coding from unstructured clinical text, such as vaccine safety reports, is a critical task with applications in pharmacovigilance and safety monitoring. Symptom coding, as tailored in this study, involves identifying and linking nuanced symptom mentions to standardized vocabularies like MedDRA, differentiating it from broader medical coding tasks. Traditional approaches to this task, which treat symptom extraction and linking as independent workflows, often fail to handle the variability and complexity of clinical narratives, especially for rare cases. Recent advancements in Large Language Models (LLMs) offer new opportunities but face challenges in achieving consistent performance. To address these issues, we propose Task as Context (TACO) Prompting, a novel framework that unifies extraction and linking tasks by embedding task-specific context into LLM prompts. Our study also introduces SYMPCODER, a human-annotated dataset derived from Vaccine Adverse Event Reporting System (VAERS) reports, and a two-stage evaluation framework to comprehensively assess both symptom linking and mention fidelity. Our comprehensive evaluation of multiple LLMs, including Llama2-chat, Jackalope-7b, GPT-3.5 Turbo, GPT-4 Turbo, and GPT-4o, demonstrates TACO's effectiveness in improving flexibility and accuracy for tailored tasks like symptom coding, paving the way for more specific coding tasks and advancing clinical text processing methodologies.
  % Accurately extracting and linking entities within clinical reports is a significant challenge in clinical data interpretation. The extraction and linking task involves identifying specific entities, such as symptoms or adverse events, within clinical reports and associating them with standardized medical codes or identifiers. Surprisingly, there has been limited exploration of Large Language Models (LLMs) for addressing this task, particularly within the Vaccine Adverse Event Reporting System (VAERS) dataset. In this study, we contribute to bridging this gap by presenting: (1) a human-annotated dataset leveraged on the VAERS dataset, specifically tailored for entity extraction and linking in clinical reports, (2) a novel evaluation scheme designed to assess LLMs' performance on this task comprehensively, and (3) an investigation into the implications of task context in prompts for enhancing extraction accuracy. Our work sheds light on the potential of LLMs in addressing this critical problem within VAERS data and offers valuable insights for future research in healthcare data processing.
\end{abstract}





\maketitle
\section{Introduction}

Accurate symptom coding from unstructured clinical text, particularly in the context of vaccine safety and pharmacovigilance, remains a critical yet complex task. Systems such as the Vaccine Adverse Event Reporting System (VAERS) \cite{VAERS} play a vital role in monitoring potential adverse events following immunization on a global scale. However, the highly variable and informal nature of clinical narratives within these reports presents significant challenges for automatically extracting and linking symptoms to standardized medical vocabularies like Medical Dictionary for Regulatory Activities (MedDRA) \cite{brown1999meddra}.

In this work, we formulate symptom coding as a task of identifying and linking nuanced symptom mentions to standardized vocabularies, which is different from broader medical coding tasks such as International Classification of Diseases (ICD) coding. While ICD coding primarily addresses diagnoses and procedures, symptom coding emphasizes the extraction of subjective experiences (e.g., ``dizziness'' or ``rash'') and their precise mapping to a predefined set of codes. This tailored task is crucial for downstream applications such as pharmacovigilance, safety monitoring, and trend analysis, offering a more flexible and specific approach compared to traditional medical coding practices.

Traditional methods for medical symptom coding typically separate symptom extraction and linking into independent workflows. Earlier approaches relied on rule-based systems or contextual models, which often struggled to capture the full context of clinical narratives and were prone to errors, especially for rare or ambiguous symptoms \cite{Gu2023DistillingLL,medcoding,tssa2023}. The disjointed nature of these processes introduced inefficiencies and inconsistencies, limiting their effectiveness in complex cases. These limitations are particularly pronounced in tasks like symptom coding, where ensuring reliable mappings of symptom mentions to standardized vocabularies is critical.

\begin{figure*}[!tp]
    \centering
    \includegraphics[trim= 0 380 0 125,clip, width=\textwidth]{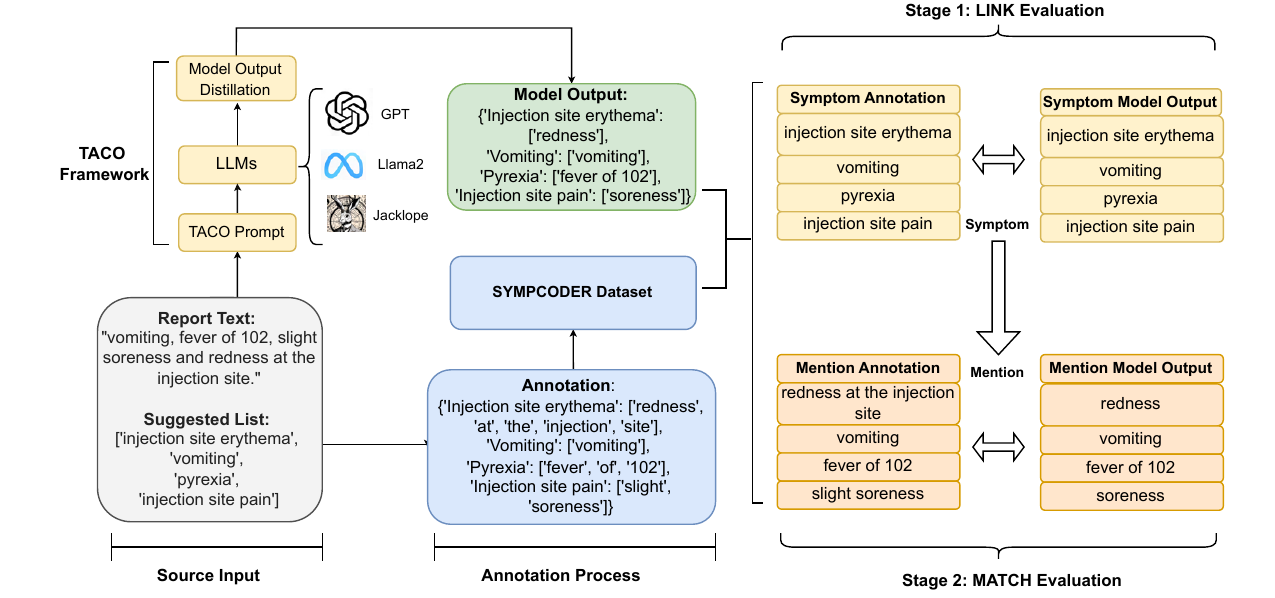}
    \vspace{-3mm}
    \caption{\ourdata\ data creation and overall workflow of TACO prompting and evaluation. Key components in the framework include: (1) Source Input: Report text and a suggested symptom list; (2) TACO Prompting: Guides LLMs in symptom coding; (3) Model Output Distillation: Refines LLM outputs; (4) SYMPCODER Dataset: Contains human annotations; and (5) Two-Stage Evaluation: LINK for matching extracted symptoms with annotations, and MATCH for assessing the contextual accuracy of symptom mentions.}
    \vspace{-3mm}
    \label{fig:general_flow}
\end{figure*}

Recent advancements in Large Language Model (LLMs) have shown promise in addressing these challenges, offering enhanced contextual understanding and nuanced language processing capabilities \cite{mahendran2021extractingadversedrugevents}. By integrating extraction and linking tasks into a single process, LLMs have the potential to overcome the limitations of traditional approaches. However, ensuring consistent accuracy across diverse cases—especially rare or complex adverse events—remains an open problem \cite{Gu2023DistillingLL}. The need for a unified framework that preserves the relationships between symptoms and their corresponding codes while maintaining efficiency is paramount.

To address these challenges, we propose the\textbf{ Task as Context Prompting (TACO)}, a novel approach leveraging LLMs to perform symptom extraction and linking as interdependent tasks. Unlike traditional methods that treat these processes in isolation \cite{medcoding}, TACO embeds task-specific context directly within LLM prompts, enabling the model to understand and maintain relationships between symptoms and standardized codes throughout the process. By unifying extraction and linking, TACO reduces information loss and enhances the flexibility and accuracy of symptom coding. 
Additionally, to the best of our knowledge, there is no existing dataset specifically designed for symptom coding in the medical domain. To support the training and evaluation of models for medical symptom coding, we introduce \textbf{\ourdata}, a human-annotated dataset derived from VAERS reports. This dataset encompasses three variants, \ourdata-Full, \ourdata-Common-50, and \ourdata-Rare-50, providing a comprehensive benchmark for assessing model performance across diverse cases, including both frequently reported and rare adverse events. By offering detailed annotations for both symptom extraction and linking, \ourdata\ enables systematic evaluation of models and facilitates further advancements in medical symptom coding research.

% \textcolor{red}{Before the evaluation, it is better to discuss the contributions of data annotation. this is a major contribution of the work. we need to highlight it in the introduction. You can start with : To the best of our knowledge, there is no existing dataset for symptom coding in the medical domain. To support the training and evaluation of models for medical symptom coding, we ...}

Our study further introduces a two-stage evaluation framework. The \textbf{Linking Integrity and Knowledge (LINK)} stage evaluates the accuracy of linking extracted symptoms to standardized codes, while the \textbf{Mention Accuracy and Textual Coherence (MATCH)} stage focuses on the fidelity and coherence of the original mentions extracted from clinical narratives. This dual-phase evaluation framework provides a granular analysis of model performance across both common and rare cases, offering comprehensive insights into their capabilities. 
Figure \ref{fig:general_flow} illustrates our TACO prompting framework, detailing the workflow from input processing to final evaluation.
In summary, our study introduces the following contributions:

\begin{itemize}[leftmargin=*] 
\item \textbf{\ourdata\ Dataset}: A human-annotated dataset based on VAERS. This dataset includes three variants and provides a benchmark to evaluate the capabilities of various methods for symptom coding, including LLMs, across diverse cases, ranging from common to rare adverse events. 

% We introduce \ourdata, a human-annotated dataset derived from VAERS reports, serving as a comprehensive benchmark for evaluating LLMs performance on adverse event extraction and linking tasks. \ourdata\ includes three variants: \ourdata-Full, \ourdata-Common-50, and \ourdata-Rare-50, enabling comprehensive assessments across diverse cases, from common to rare adverse events.

\item \textbf{\ourprompt\ Prompting Framework}: A novel approach that unifies symptom extraction and linking by embedding task-specific context within prompts, improving flexibility and accuracy.

% We propose the TACO framework, which integrates task-specific context directly within LLM prompts, enhancing the flexibility and accuracy of symptom coding by unifying extraction and linking processes. 

\item \textbf{Comprehensive Two-Stage Evaluation}: A dual-phase framework (LINK and MATCH) that assesses both the linking accuracy of extracted symptoms and the coherence of their original mentions, providing a detailed understanding of model capabilities.

% \textcolor{red}{mention the two stage evaluation first.} We conduct an extensive evaluation of state-of-the-art LLMs, including open-source models like Llama2-chat \cite{llama2} and Jackalope-7b \cite{jackalope}, as well as proprietary models like GPT-3.5 Turbo \cite{gpt3.5}, GPT-4 Turbo \cite{gpt4turbo}, and GPT-4o \cite{gpt4o}. Our study compares sequential and integrated prompting methods, assessing their effectiveness in enhancing the extraction and linking of adverse events. 

\end{itemize}

\section{Related Work}
\subsection{Entity Extraction and Linking}
Entity extraction and linking are fundamental tasks in natural language processing (NLP) that facilitate the transformation of unstructured text into structured, actionable data. These tasks are widely used in domains such as general text processing, biomedical information retrieval, and clinical text mining \cite{JEHANGIR2023100017}.

Entity extraction focuses on identifying relevant entities (e.g., people, places, symptoms) in text. Earlier approaches to this task relied on rule-based systems and statistical models, such as Hidden Markov Models (HMMs) \cite{baum1966statistical} and Conditional Random Fields (CRFs) \cite{PATIL20201181}. While these models were effective in specific scenarios, they struggled with generalizing across diverse contexts and handling the inherent variability of natural language \cite{keraghel2024surveyrecentadvancesnamed}. The introduction of deep learning, particularly Recurrent Neural Networks (RNNs) and Long Short-Term Memory (LSTM) networks, improved the ability to capture sequential dependencies and contextual nuances \cite{Ali2022NamedER}. Recently, transformer-based models such as BERT \cite{devlin-etal-2019-bert} have revolutionized entity extraction by leveraging self-attention mechanisms, enabling the capture of long-range dependencies and complex contextual relationships \cite{qiao2022joint}.

Entity linking complements extraction by resolving ambiguities and connecting extracted entities to predefined ontologies or knowledge bases such as Wikipedia \cite{wikidump}, DBpedia \cite{auer2007dbpedia}, or medical databases like SNOMED \cite{jama.1980.03300340032015} and UMLS \cite{bodenreider2004unified}. Traditional methods typically employed a two-step approach: extracting handcrafted features and using entity-ranking models for linking predictions. These methods were limited by their reliance on labor-intensive feature engineering and poor generalizability across different KBs and domains \cite{eldl}. Modern approaches such as embedding-based method, powered by models like BERT, have significantly enhanced linking by mapping textual entities and knowledge base entries into a shared vector space \cite{broscheit2020investigating}. Despite advancements, embedding-based methods still face challenges such as reliance on large annotated datasets, difficulty generalizing to unseen entities or domains, and computational inefficiency at scale, while often struggling with nuanced contextual relationships in specialized domains \cite{eldl}.

In the biomedical domain, extraction and linking tasks are often addressed independently, overlooking their inherent interdependencies and potential benefits of integrating them. However, combining both extraction and linking tasks is critical for applications like ICD coding and adverse event detection, where seamless integration not only enhances the accuracy but also improves the overall efficiency by providing essential contextual information to each other \cite{kartchner2023bioel,french2023overview}. 
BioBERT-based Named Entity Recognition and Normalization (BERN) \cite{bern} marked a milestone in this field by unifying entity extraction and linking into a single, streamlined framework. Leveraging a fine-tuned version of BioBERT \cite{biobert}, BERN integrates these interdependent tasks and eliminates the need for separate modules, achieving state-of-the-art performance at the time. This integrated approach laid the groundwork for further advancements in medical coding. However, significant gaps and challenges remain, highlighting the need for further investigation.

\subsection{Advances in Contextual Models and LLMs}
Contextual language models like BERT and its domain-specific adaptations, such as BioBERT and ClinicalBERT \cite{clinicalbert}, have significantly improved tasks involving complex biomedical text. By incorporating domain-specific pretraining on large biomedical corpora, these models enhanced the accuracy of tasks like symptom extraction, diagnosis identification, and treatment mapping \cite{mahendran2021extractingadversedrugevents,ji2021does}. Despite these advances, they continued to treat extraction and linking as distinct tasks, requiring additional post-processing steps that often introduced errors, particularly for ambiguous or infrequent cases \cite{medcoding}.
Moving beyond these limitations, CNN-based architectures such as MultiResCNN \cite{multirescnn} further advanced medical coding by incorporating residual blocks and multi-resolution filters to capture complex patterns in clinical notes. However, these models required explicit alignment of input features with predefined codes, limiting their adaptability to unseen data or emerging clinical terms.

The emergence of large language models (LLMs) like GPT-3 and GPT-4 has transformed NLP, enabling the integration of extraction and linking tasks within a single framework. For example, LLM-Codex \cite{codex} utilized a two-stage pipeline to improve the reliability of ICD coding predictions, with an LLM in the first stage and a verifier model in the second stage. Similarly, Boyle et al. \cite{boyle2023automated} proposed a zero-shot and few-shot ICD coding method using pre-trained LLMs, framing the task as an information extraction problem and leveraging the hierarchical structure of the ICD ontology for efficient code assignment. While these approaches demonstrated state-of-the-art performance, their reliance on predefined hierarchies and focus on ICD coding distinguishes them from our work.

\section{The \ourdata \ Dataset Creation}

% \textcolor{red}{please organize this section with the following 3 subsections.}

\subsection{VAERS Dataset}

The Vaccine Adverse Event Reporting System (VAERS) dataset, managed by the Centers for Disease Control and Prevention (CDC) and the Food and Drug Administration (FDA), is a critical resource for monitoring vaccine safety. VAERS operates as a passive reporting system where individuals, including healthcare professionals and the general public, can submit reports of adverse events following immunization. Healthcare professionals must report certain adverse events, and vaccine manufacturers must report all adverse events they become aware of. VAERS does not determine causality but helps detect unusual or unexpected patterns in adverse event reports that may suggest safety issues, aiding the CDC and FDA in identifying areas that need more evaluation and assessment. Since 1990, VAERS has collected millions of reports, each containing detailed information such as demographics, vaccination details, descriptions of adverse events, standard symptom lists, and medical history \cite{VAERS}. This extensive dataset serves as a valuable resource for medical symptom coding, facilitating ongoing research and analysis to enhance our understanding of vaccine safety and address emerging concerns.

\begin{table}[!tp]
    \centering
    \caption{Basic Data Statistics of \ourdata. 
    Note that ``Suggested Symptoms'' refers to the labeled symptoms in the SYMPCODER dataset 
(not the ``suggested list'' from VAERS), and “Extracted Symptoms” are those automatically 
identified by LLMs.}
    \vspace{-2mm}
    \resizebox{1.0\linewidth}{!}{
    \begin{tabular}{lccc}
    % { p{4cm}p{2cm}p{2cm}p{2cm}}
    \toprule
     & \bf Clinical Text & \bf Suggested Symptoms & \bf Extracted Symptoms\\
     \midrule
     \multicolumn{4}{l}{\textbf{\ourdata-Full} (\# of Reports: 487)} \\
     Average Length & 843 & 8 & 6 \\
     Median Length & 321 & 5 & 5 \\
     Min Length & 9 & 1 & 0 \\
     Max Length & 11834 & 105 & 41 \\
     \midrule
     \multicolumn{4}{l}{\textbf{\ourdata-Common-50} (\# of Reports: 427)}  \\
     Average Length & 873 & 9 & 6 \\
     Median Length & 326 & 6 & 5 \\
     Min Length & 9 & 1 & 0 \\
     Max Length & 11834 & 105 & 41 \\
     \midrule
     \multicolumn{4}{l}{\textbf{\ourdata-Rare-50} (\# of Reports: 22)}  \\
     Average Length & 1659 & 9 & 8 \\
     Median Length & 1000 & 8 & 7 \\
     Min Length & 30 & 2 & 0 \\
     Max Length & 8555 & 30 & 21 \\
     \bottomrule
    \end{tabular}
    }
    
\label{tab:data_statistics}
% \vspace{-5mm}
\end{table}

\subsection{Entity Mention Annotation}
% For our study, a sample of 500 VAERS reports spanning from 1990 to 2023 was randomly selected. To ensure accuracy and reliability, a team comprising three annotators and one validator was assembled. Each VAERS report is accompanied by a list of suggested symptom terms, encoded according to MedDRA terminology. These suggested terms indicate potential adverse events based on the text description provided in each report. The purpose of annotation was to identify and mark all mentions of symptoms explicitly related to vaccine adverse events for each suggested symptom term. By capturing these mentions, we ensure the dataset reflects the full variability of how symptoms are expressed, which is critical for creating a robust dataset to train and evaluate models on medical symptom coding. 

We randomly selected 500 VAERS reports (1990–2023) and assembled a team of three annotators plus one validator to ensure accuracy. Each report included suggested symptom terms (MedDRA-encoded), and our annotation process marked all symptom mentions related to vaccine adverse events. This approach captures the variability of symptom expression and yields a robust dataset for model training and evaluation.
The annotation process consists of two key phases, including \textbf{annotation} and \textbf{validation}, designed to ensure both thoroughness and accuracy.

\subsubsection{Annotation Phase}
% The annotation phase was carried out meticulously by a team of three graduate students selected based on their academic background, technical proficiency in data annotation, familiarity with the English language, and basic clinical knowledge. These criteria ensured that the team could handle the linguistic and domain-specific nuances required for accurate annotation. 
The annotation phase was meticulously carried out by three graduate students selected for their academic background, annotation skills, English proficiency, and basic clinical knowledge, ensuring they could handle the linguistic and domain-specific nuances required for accurate annotation.
Each annotator was responsible for annotating approximately one-third of the total 500 reports for annotations by following a clearly defined set of guidelines to ensure consistency and reliability across all reports.
During the annotation, (1) the annotators used a home-maintained data annotation tool designed specifically for this task. This tool ensured a seamless workflow, focusing solely on adverse events while disregarding procedural details and negative test results to eliminate any irrelevant information.
(2) Each report was annotated one label at a time, using a list of suggested terms for symptoms provided in the original VAERS data, which are encoded according to the MedDRA terminology. This step-by-step process allowed the annotators to focus on one potential adverse event at a time, ensuring accuracy and precision.
(3) In cases where annotators encountered uncertainties or ambiguities in the text, they marked the annotation as "uncertain" for further validation. This precautionary measure allowed a second layer of scrutiny to improve the reliability of the dataset.

\subsubsection{Validation Phase}
The validation phase was conducted by a senior annotator specializing in natural language processing for the medical domain, selected for his or her extensive annotation experience. 
(1) The validator meticulously reviewed all annotations, especially those marked as "uncertain", to ensure flagged annotations received additional scrutiny and maintained high data integrity.
(2) For each uncertain annotation, the validator engaged in discussions with the original annotator to understand their reasoning and reach a consensus. Persistent ambiguities were resolved collaboratively within the team to ensure alignment with the annotation guidelines.
(3) When discrepancies or inconsistencies were identified, the validator suggested revisions, which were reviewed and discussed with the annotators. Final decisions on adjustments were made collaboratively to ensure consistency and transparency throughout the validation process.

The resulting human-annotated dataset, \textbf{\ourdata}, stands as a fundamental benchmark for medical symptom coding tasks. Beyond its utility in our study, this dataset offers valuable insights for advancing research in this field and exploring the capabilities of various methods for symptom coding in future investigations. This rigorous annotation and validation process ensured high data integrity and reliability, making \ourdata\ a robust resource for evaluating the performance of LLMs in the symptom coding task presented in this study.

\subsubsection{\ourdata\ Dataset}
We constructed three subsets of the annotated dataset: \ourdata-Full, \ourdata-Common-50, and \ourdata-Rare-50 to facilitate focused analyses. These subsets were created to evaluate model performance across symptoms with different frequencies, targeting variations between common cases and rare or ambiguous cases. 
Additional details about the dataset and project can be found in \url{https://github.com/LEAF-Lab-Stevens/TACO-Prompting}.
\begin{itemize}[leftmargin=*] 
    \item \textbf{\ourdata-Full}: This comprehensive dataset includes all annotated symptom mentions from the VAERS reports, providing a holistic benchmark for symptom extraction and linking tasks.
    \item \textbf{\ourdata-Common-50}: This subset focuses on the 50 most frequently mentioned symptoms in the dataset. Symptoms were ranked by their frequency of occurrence within \ourdata-Full. The top 50 symptoms were selected, and reports containing at least one of these symptoms were included in this subset.
    \item \textbf{\ourdata-Rare-50}: This subset highlights the 50 least frequently mentioned symptoms. Following a similar process to the Common-50 subset, symptoms were ranked by frequency, and reports containing at least one of these rare symptoms were included in this subset.
\end{itemize}

\subsection{Basic Statistics of \ourdata}

% Prompting Method Example,notes: combine bins by grouping
% length or distribution
% more plots to show dataset
% top 50 vs bot 50 with diff color in the same plot
% Or one for each

Of the initial 500 annotations, 487 distinct reports remained after removing 23 uncertain results, which annotators and validator could not verify as symptoms due to limited domain knowledge or complex contexts. The \ourdata\ datasets, comprising \ourdatafull, \ourdatacom, and \ourdatarare, provide a robust foundation for further studies. As shown in Table \ref{tab:data_statistics}, the \ourdatafull\ dataset offers a comprehensive resource, capturing diverse adverse event descriptions with varying lengths and complexities. Table \ref{tab:data_statistics} provides additional details, such as the average and median lengths of clinical text, suggested symptoms, and extracted symptoms. These metrics reveal consistent gaps between human-labeled and model-generated symptoms across all subsets, highlighting the challenges of achieving complete extraction. By focusing on common and rare subsets, the \ourdatacom\ and \ourdatarare\ datasets provide targeted benchmarks for evaluating model performance across both frequent and infrequent adverse events.

\begin{figure*}[!tp]
\centering
	\begin{subfigure}[b]{0.3\textwidth}
		\includegraphics[width=\textwidth]{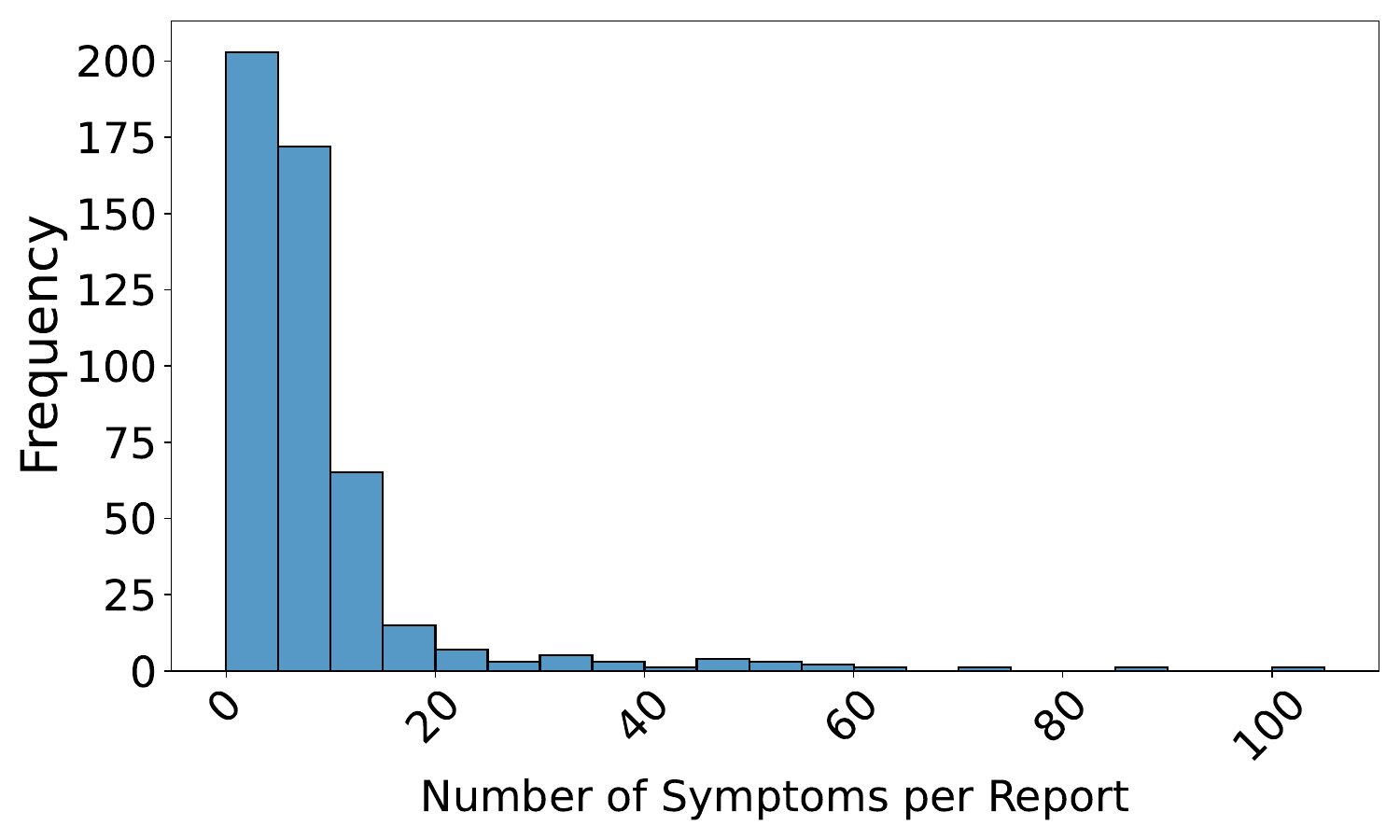}
		\vspace{-5mm}
		\caption{\ourdatafull}
		\label{fig:dist_full}
	\end{subfigure}
        \quad
	\begin{subfigure}[b]{0.3\textwidth}
		\includegraphics[width=\textwidth]{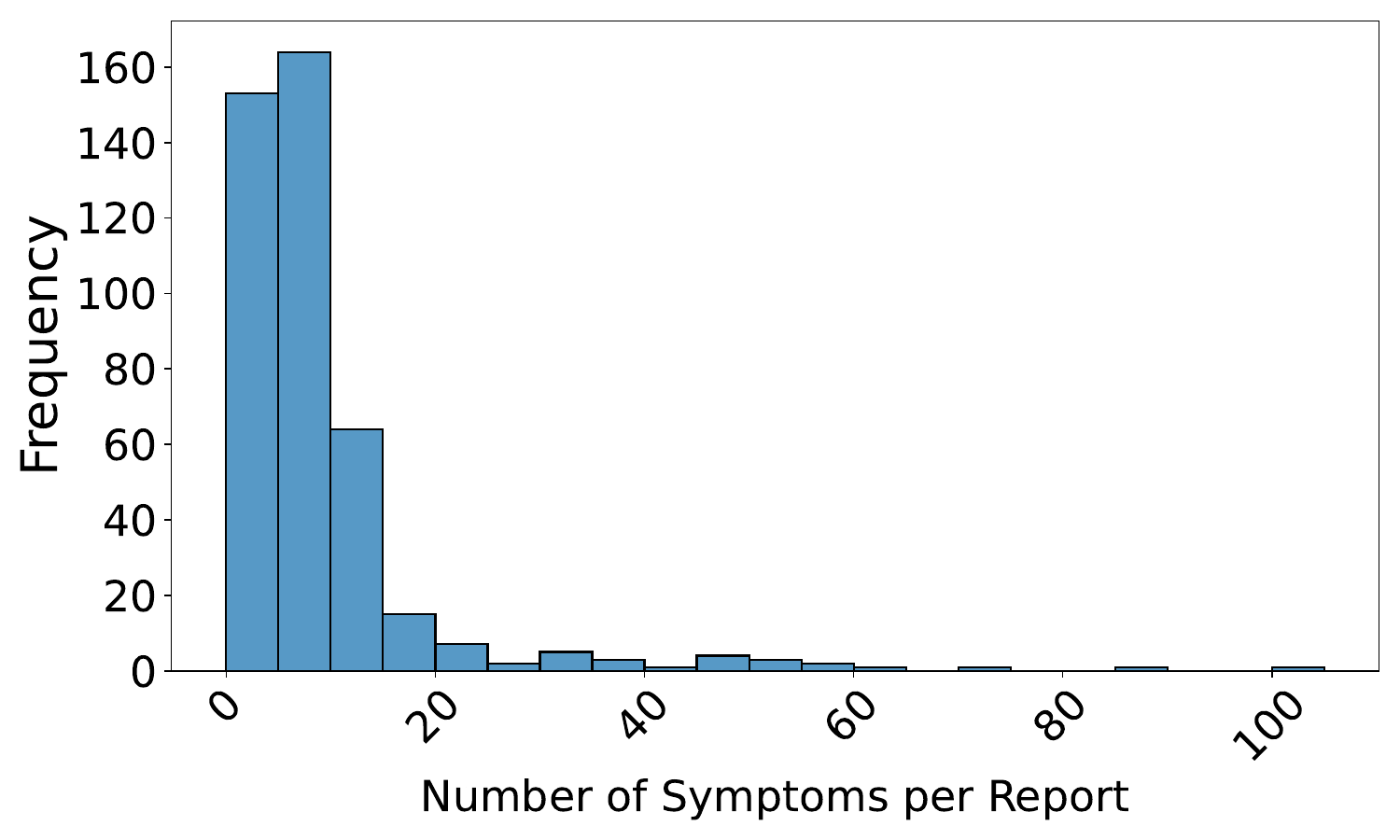}
		\vspace{-5mm}
		\caption{\ourdatacom}
		\label{fig:dist_com}
	\end{subfigure}
        \quad
        \begin{subfigure}[b]{0.3\textwidth}
		\includegraphics[width=\textwidth]{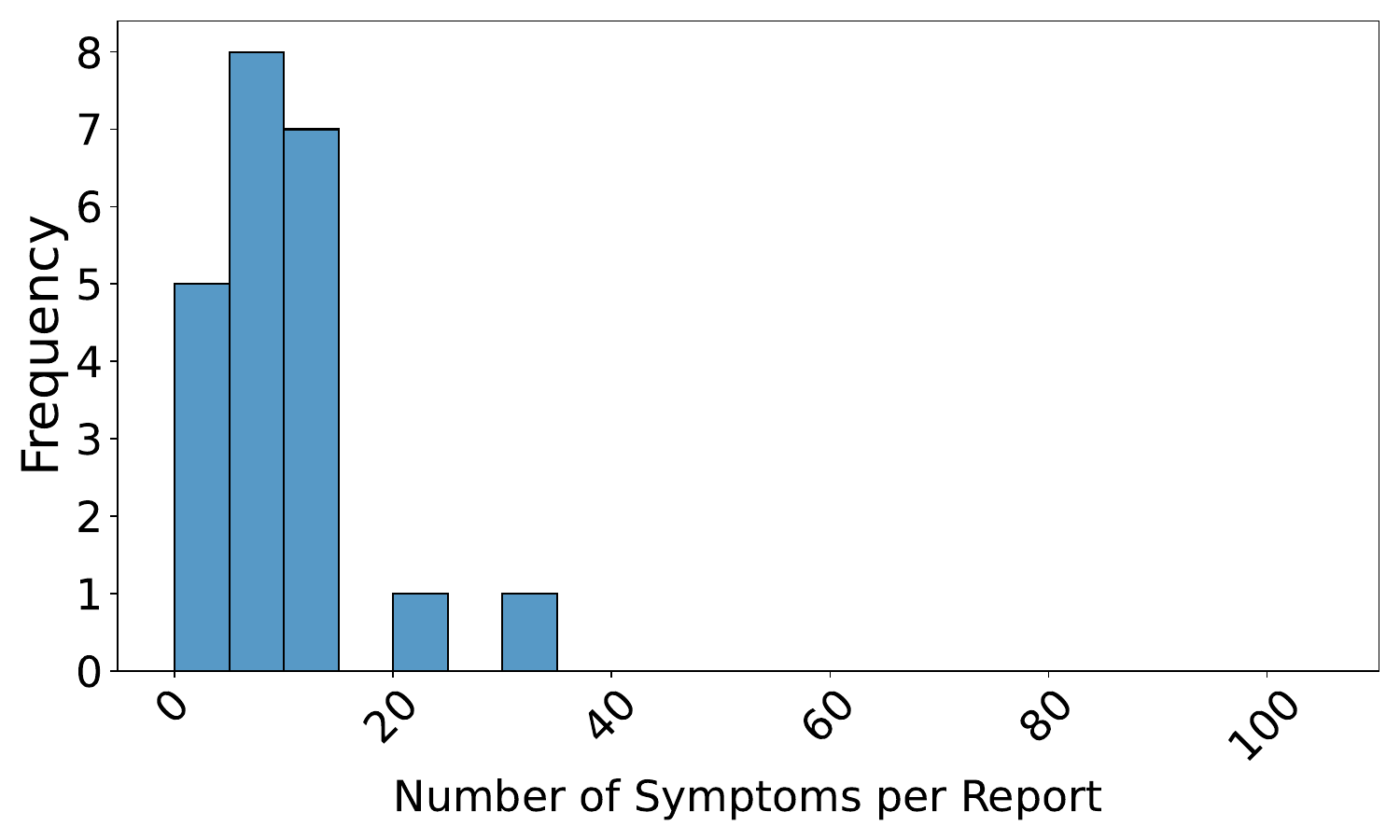}
		\vspace{-5mm}
		\caption{\ourdatarare}
		\label{fig:dist_rare}
	\end{subfigure}
	
	% \vspace{-4mm}
	% \caption{Common Rare Case Analysis with TASI Prompting on \ourdata }
	\vspace{-3mm}
	\label{fig:dataset analysis}

    \caption{Distributions of the number of symptoms for different datasets in \ourdata.}
\label{fig:symp_freq}
\end{figure*}

Figure \ref{fig:dist_full} shows that most reports in the \ourdatafull\ dataset contain between 0 and 10 symptoms, with a sharp decline for higher counts. A similar trend is observed in Figure \ref{fig:dist_com} for the \ourdatacom\ dataset. Figure \ref{fig:dist_rare} highlights a steeper drop after 10 symptoms in the \ourdatarare\ dataset, with a few outliers reaching up to 40 symptoms, reflexting the rarity and complexity of such cases.

% \yue{I think table 1 summarizes these data statistics very well. Do we need to describe these details in text again?}
% The \ourdatafull\ dataset includes all annotated adverse events, offering a broad spectrum for evaluation. As shown in Table \ref{tab:data_statistics}, the length of adverse event descriptions ranges from 9 to 11,834 characters, with a median of 321 and an average of 843 characters. The number of suggested terms per report varies from 1 to 105, averaging 8 terms, while model-extracted terms per report range from 0 to 41, with an average of 6. This consistent gap between suggested and extracted terms underscores the challenges of comprehensive extraction, particularly in complex cases.

% The \ourdatacom\ dataset consists of 427 reports focusing on the 50 most frequently mentioned symptoms. The length of adverse event descriptions ranges from 9 to 11,834 characters, with a median of 326 and an average of 873 characters. The number of suggested terms per report averages 8, while model-extracted terms average 6. These metrics mirror those of the full dataset, reflecting the difficulty of accurately capturing common symptoms.

% The \ourdatarare\ dataset contains 22 reports centered on the 50 least frequently mentioned symptoms. Adverse event descriptions range from 30 to 8,555 characters, with an average length of 1,659 characters. Suggested terms per report average 9, while model-extracted terms average 8. The smaller gap between suggested and extracted terms suggests that rare symptoms are less complex and more targeted for extraction.

\begin{figure}[!tp]{
\includegraphics[trim = 15 10 0 20,clip, width=0.5\textwidth]{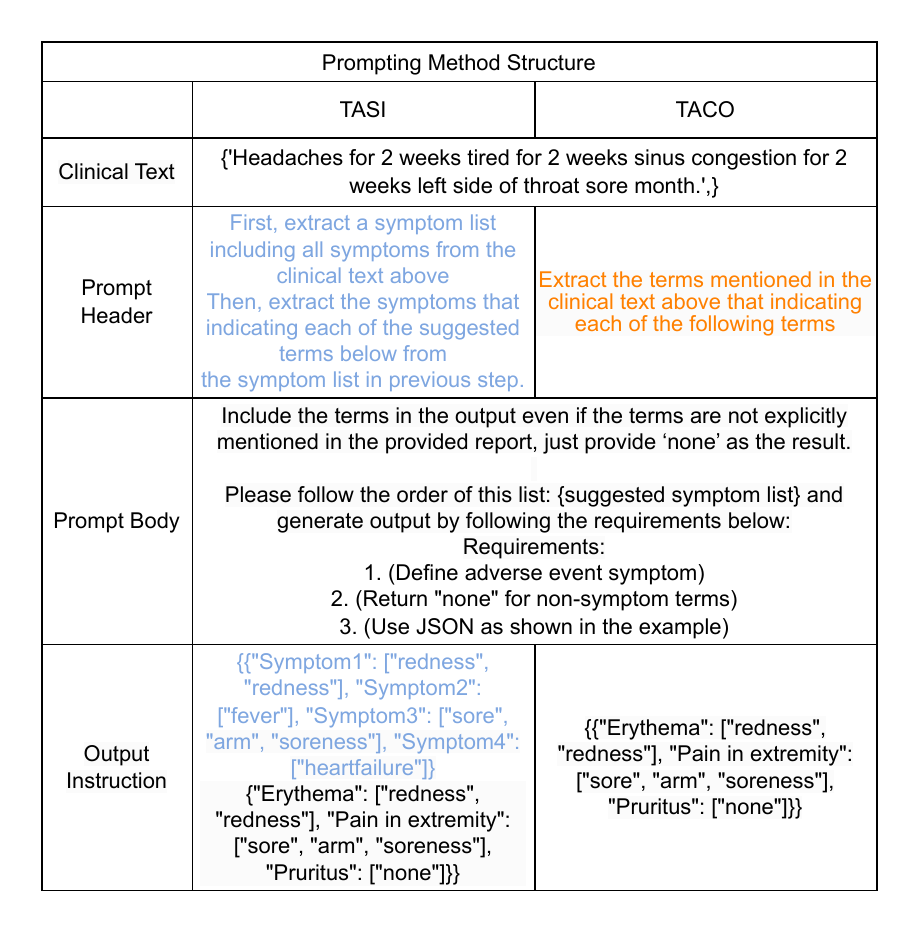}
\vspace{-3mm}
\caption{The structure of TASI and TACO prompts, detailing clinical input, task instructions, and output format. The provided output examples are for format demonstration purposes only and do not align with the clinical input text, which is taken from real clinical reports.}
\label{fig:prompt variation}}
\vspace{-3mm}
\end{figure}

\section{The Proposed \ourprompt\ Prompting}
% \yue{should we change the section title to be ``The Proposed TACO framework''?}
% \textcolor{red}{try to organize this section into 4.1-4.3. 
% Sequential Integrated Prompting Analysis/Evaluation/Retrieval Method}
% Need more stats and well defined names

% The primary objective of this paper is to evaluate the efficacy of Large Language Models (LLMs) \textcolor{red}{we have already provided this abbreviation in previous sections, we can directly use LLMs here} in extracting and linking vaccine adverse events from clinical reports to a predefined list of potential adverse event terms. "Linking" in this context refers to the process of matching terms mentioned in the original medical reports to the formal MedDRA codes provided in the suggested list for each report. To assess the accuracy of LLMs in performing this extraction and linking task, we designed \textbf{Ta}sk as \textbf{Co}ntext (\ourprompt) \textcolor{red}{put `prompting' outside} method tailored for this purpose. 

This paper evaluates the efficacy of various LLMs in extracting vaccine adverse event symptoms from VAERS reports and linking to formal MedDRA codes in a predefined list. To address this task, we designed \textbf{T}ask \textbf{a}s \textbf{Co}ntext (\ourprompt) prompting, a novel prompting method that integrates symptom extraction and linking into a unified framework. By embedding the interrelated tasks within a single prompt, \ourprompt\ leverages the broader task context to enhance model performance and adaptability for automated symptom coding. To benchmark its effectiveness, we also designed \textbf{T}ask \textbf{a}s \textbf{S}equent\textbf{i}al (\alterprompt), which separates both tasks into distinct phases, allowing for a comparative analysis of the two approaches. Throught this comparison, our study highlights how embedding interrelated tasks, as proposed by the Task as Context strategy, enhances the performance and adaptability of LLMs for automated medical symptom coding.

\subsection{Task as Context Prompting}
% \yue{Maybe this subsection can be ``Task as Context Prompting''?}
% \textcolor{red}{this subsection is too long, making it difficult to understand. suggestions: add the preliminaries section to introduce the task as context idea, see more details below. then in the prompt design subsection, you focus on introducing more details of the taco prompting, and clearly list the corresponding components, }

\subsubsection{\textbf{Inspiration}}
% \textcolor{red}{first introduce the task as context idea in the hcomp paper, what task it is designed for, what is the motivation, and benefits of this strategy. then, introduce why we adopt this strategy for the prompting design in our work: because we have two highly related and interdependent tasks, including extraction and linking, traditionally, these two tasks are solved independently, leading to the limitations, mention the limitations; what is the advantage of addressing these two tasks together; then how the task as context strategy can help us achieve this goal. therefore, in this work, we design taco prompting to ...}
The Task as Context concept, originally designed for annotation workflows with non-experts involving interdependent tasks, emphasizes leveraging contextual relationships to improve accuracy and adaptability. This strategy integrates related tasks into a single process, reducing inefficiencies and error propagation \cite{li2023task}. In the context of symptom coding, two highly interdependent tasks, symptom extraction and linking symptoms to standardized codes, are traditionally solved independently. However, this modular approach suffers from limitations, including the loss of contextual cues between tasks, error propagation from one task to another, and scalability challenges due to the need for separate training processes.

Inspired by the Task as Context strategy, we address these limitations by designing \ourprompt\ prompting, which explicitly embeds the interdependence of extraction and linking into a unified prompting framework. \ourprompt\ integrates the tasks within the prompt, enabling simultaneous learning and execution. By treating the two tasks holistically, our approach improves contextual understanding, minimizes error propagation, and enhances both accuracy and scalability, providing a streamlined and adaptive solution for symptom coding challenges.

% which embeds task-specific context into LLM prompts. This approach unifies symptom extraction and linking within a single framework, allowing the model to leverage inter-task dependencies effectively. 

% % \yue{I feel this paragraph can be moved to a later section to introduce the overall framework like section 4.1} 
% In contrast, our work introduces \ourprompt\ prompting, which explicitly embeds the interdependence of extraction and linking into a unified prompting framework. Unlike LLM-Codex, which applies post hoc verification, TACO integrates the tasks within the prompt, enabling simultaneous learning and execution. This approach ensures improved consistency across sentence- and document-level coding tasks, making it more adaptable to varied clinical datasets and challenging scenarios involving rare symptoms. By leveraging TACO, we demonstrate the potential for achieving superior accuracy and robustness compared to existing methods.
\subsubsection{\textbf{\ourprompt\ Prompting Design}}

To clearly present the structure of TACO prompting, we outline its four core components in Figure \ref{fig:prompt variation}.
\begin{itemize}[leftmargin=*] 
    \item \textbf{Clinical Text:}  
    The clinical text serves as the raw, unstructured input data from the VAERS reports. It mirrors real-world scenarios where models must process free-text narratives to extract and link relevant symptoms. This component provides the context required for understanding the input data and serves as the foundation for subsequent tasks.

    \item \textbf{Prompt Header:}  
    The prompt header provides explicit instructions for task execution. In TACO, this component integrates both symptom extraction and linking tasks into a single step. It instructs the model to directly extract symptoms from the clinical text and map them to the provided suggested terms, leveraging the broader task context. In contrast, the benchmark prompt TASI separates these tasks into two distinct phases, requiring the model to first extract all symptoms and subsequently link them to suggested terms.

    \item \textbf{Prompt Body:}  
    The prompt body contains detailed guidelines for task execution and ensures that both tasks are carried out as intended. For TACO, the instructions guide the model to simultaneously address symptom extraction and linking, thereby reducing redundancy and improving efficiency. TASI, however, employs sequential instructions, where symptom extraction is followed by linking. This distinction reflects the core advantage of TACO, which embeds the broader task context into a cohesive framework to improve task understanding and execution.

    \item \textbf{Output Instruction:}  
    The output instruction specifies the expected output format to ensure consistency and clarity. Both TACO and TASI utilize a structured JSON format; however, their approaches differ. In TACO, the format unifies the output by directly mapping symptoms to suggested terms, aligning with its integrated task-solving approach. In TASI, the output separates extraction and linking results, reflecting its sequential methodology. The inclusion of illustrative examples further clarifies the expected output structure for the model.
\end{itemize}

To comprehensively evaluate the effectiveness of TACO, we compare it against the benchmark prompt TASI, which adheres to traditional sequential task-solving methodologies. While TASI serves as a baseline, TACO’s integrated and context-rich design seeks to overcome the limitations of traditional approaches by streamlining task execution and enhancing contextual understanding.

\subsubsection{\textbf{Model Output Distillation and Evaluation Tasks}}
\label{eval-tasks}

To address inconsistencies in model outputs, we incorporate a post-processing pipeline during the model output distillation phase, leveraging Regular Expression (regex) \cite{regex} syntax to systematically capture the information needed. These inconsistencies include incomplete model outputs, unnecessary descriptive text at the beginning of responses, and nonsensical outputs that deviate from task requirements. The distillation process ensures uniformity and precise extraction of relevant information from the varied responses generated by LLMs. As shown in Figure \ref{fig:general_flow}, the \ourprompt\ outputs undergo this refinement step, eliminating irrelevant details and preparing them for structured evaluation.

The evaluation phase comprises two stages. In the first stage, known as \textbf{L}inking \textbf{In}tegrity and \textbf{K}nowledge (LINK), we assess the models based on how accurately they link extracted symptoms to the corresponding terms from the Suggested List. In this stage, as Figure \ref{fig:general_flow} shows, the model must first identify the correct symptoms (e.g., "injection site erythema," "vomiting") from the clinical text using the distilled output from the previous phase. The focus here is on ensuring that each relevant symptom has been accurately identified and linked to the correct terminology. This stage evaluates the model's ability to extract and link adverse events, providing insights into how well it handles clinical information within the context of symptom extraction.

In the second stage, termed \textbf{M}ention \textbf{A}ccuracy and \textbf{T}extual \textbf{C}o\textbf{h}erence (MATCH), the focus shifts to the quality of the original mentions generated by the model for each linked symptom. As depicted in Figure \ref{fig:general_flow}, this involves comparing the generated model results to the human-annotated gold standard in terms of specificity and precision. For example, while LINK verifies if the model has correctly linked "injection site erythema," MATCH evaluates whether the model-generated mention—such as "redness at the injection site"—is semantically similar to the original clinical report, which is annotated simply as "redness." This two-tiered evaluation provides a deeper analysis of how well the models not only identify but retain the original context and details from the clinical text.

\subsection{Benchmark LLMs}

We aim to harness the capabilities of several cutting-edge Large Language Models (LLMs) for the extraction and linking of adverse events on the \ourdata\ dataset. Each model offers unique strengths and characteristics, making them suitable for various aspects of this complex task.

% \hl{itemize, and add citations for each model}
\begin{itemize}[leftmargin=*]
    
\item \textbf{Jackalope-7b} \cite{jackalope}: The smallest open-sourced model in our investigation, Jackalope-7b is fine-tuned by SlimOrca using several open datasets on top of Mistral 7B. Despite its smaller size, Jackalope-7b is optimized for specific NLP tasks and demonstrates a balanced trade-off between performance and computational efficiency. Its architecture allows for quicker adjustments and fine-tuning, making it particularly effective in scenarios requiring high precision within a smaller model footprint.

\item \textbf{Llama2-13b-chat} \cite{llama2}: Developed by Meta AI, Llama2-13b-chat is a medium-sized model based on the widely recognized Llama2. Known for its advanced natural language understanding capabilities, Llama2-13b-chat is designed to handle complex language tasks with a focus on generating human-like responses. Its medium size strikes a balance between computational demand and performance, making it a versatile choice for a variety of NLP applications.

\item \textbf{GPT-3.5-Turbo} \cite{gpt3.5}: An efficient variant of OpenAI's GPT-3, GPT-3.5-Turbo is designed to excel in natural language processing tasks with optimized performance. It leverages the strengths of GPT-3 while incorporating enhancements that improve response coherence, accuracy, and efficiency. This model is well-suited for applications requiring robust language comprehension and generation capabilities.

\item \textbf{GPT-4-Turbo} \cite{gpt4turbo}: An enhanced version of GPT-3.5-Turbo, GPT-4-Turbo offers comprehensive improvements in terms of model architecture, training data diversity, and computational efficiency. It is designed to handle more complex language tasks with greater accuracy and faster response times. GPT-4-Turbo's advancements make it a powerful tool for sophisticated NLP applications, including detailed extraction and linking tasks.

\item \textbf{GPT-4o} \cite{gpt4o}: A faster and more cost-effective variant of GPT-4-Turbo, GPT-4o provides efficient performance at a reduced cost. It retains many of the advanced features of GPT-4-Turbo while optimizing for speed and resource utilization. GPT-4o is particularly advantageous in environments where computational resources are limited, but high performance is still required. Its ability to balance cost and efficiency makes it a practical choice for extensive NLP tasks.

\end{itemize}
% We aim to harness the capabilities of several cutting-edge LLMs for the extraction and linking of adverse events on the \ourdata\ dataset. Jackalope, the smallest open-sourced model in our investigation, is fine-tuned by SlimOrca and several open datasets on Mistral 7B. Llama2-13b-chat, a medium-sized model based on Meta AI’s Llama2, is recognized for its advanced natural language understanding. GPT-3.5 Turbo, an efficient variant of OpenAI's GPT-3, excels in natural language processing. GPT4-Turbo, an enhanced version of GPT-3.5 Turbo, offers comprehensive improvements. GPT4o, a faster and cost-effective variant of GPT4-Turbo, provides efficient performance at a reduced cost.

% These models are esteemed for their prowess in natural language understanding. Each model was assigned the same task of extracting and linking adverse events from VAERS reports based on the given list of suggested terms, guided by our tailored input prompts. The results generated by these models will be systematically analyzed and compared against our human-annotated ground truth dataset. This comparative analysis serves as the foundation for evaluating the performance of each LLM in capturing adverse event entities within the VAERS reports.

% \subsection{Human Annotation}

% \subsection{LLM Annotation}

% \subsection{Innovative Evaluation Method with Post-Processing}

% too many emphasized values are kinda distracting. Maybe just keep the best and second bests results.

\begin{table*}[!tp]
    \centering
    \vspace{-3mm}
    
    \caption{LINK stage results assessing the accuracy of linking extracted symptoms to standard medical codes. Bold values indicate the best performance for each prompting method, while italic values show the overall best performance across methods.}
    \label{tab:first_stage_eval}
    
    \resizebox{\textwidth}{!}{
    \begin{threeparttable}

    \begin{tabular}{
    p{1.2cm}p{3cm}p{1.5cm}p{1.5cm}p{1.5cm}p{1.5cm} p{1.5cm}p{1.5cm}  }\toprule
     \bf Prompt Type& \bf Models& \bf EM-Precision & \bf EM-Recall & \bf Fuzzy-Precision & \bf Fuzzy-Recall&   \bf EM-Fuzzy-Precision &  \bf EM-Fuzzy-Recall\\
     \midrule
     & Jackalope-7b&0.874  &0.841 & 0.876 & 0.843 & 0.877 & 0.844 \\
     & Llama-2-13b-chat&0.765  & 0.582  & 0.769  & 0.584 & 0.772 & 0.585 \\
     TASI& gpt-3.5-turbo& 0.838 & 0.814 & 0.846 & 0.821 & 0.853 &0.827 \\
     & gpt-4-turbo& 0.887 & 0.867 & 0.892 & 0.872 & 0.895 &0.875 \\
     & gpt-4o& \textbf{0.912} & \textbf{0.896} & \textbf{0.918} & \textbf{0.899} & \textbf{0.921} &\textbf{0.902} \\
     \midrule
     & Jackalope-7b&0.763  &0.827 & 0.767 & 0.831 & 0.774 & 0.840 \\
     & Llama-2-13b-chat&0.886  & 0.875  & 0.893  & 0.881 & 0.902 & 0.891 \\
     TACO& gpt-3.5-turbo& 0.844 & 0.867 & 0.844 & 0.867 & 0.847 &0.869 \\
     & gpt-4-turbo& \textit{\textbf{0.999}} & \textit{\textbf{0.998}} & \textit{\textbf{0.999}} & \textit{\textbf{0.998}} & \textbf{0.999} &\textit{\textbf{0.998}}\ \\
     & gpt-4o& 0.995 & 0.994 & 0.997 & 0.995 & 0.999 &0.996 \\
     
    \bottomrule
    \end{tabular}
    \end{threeparttable}
    }

\end{table*}

\section{Experiments}
Based on the \ourdata\ dataset,
three research questions (RQs) are studied in this paper, including 
% \textcolor{red}{provide the research questions based on the experiments we have}
  
% (1) overall performance of LLMs on the task
% (2) differences between two prompts
% (3) performance on extracting the top 50 and bottom 50
% \begin{enumerate}[leftmargin=*]
%     \item \textbf{RQ1}: How do different LLMs perform in the task of extracting and linking vaccine adverse events?
%     \item \textbf{RQ2}: How do TASI and TACO prompting impact the performance of LLMs on our task?
%     \item \textbf{RQ3}: How do LLMs perform in extracting and linking the top 50 most common adverse events compared to the bottom 50?
% \end{enumerate}

\begin{enumerate}[leftmargin=*]
    \item \textbf{RQ1}: How do the TACO and TASI prompting strategies impact the performance of LLMs in the symptom coding task?
    \item \textbf{RQ2}: How do different LLMs perform on the symptom coding task when evaluated with the same prompt design?
    \item \textbf{RQ3}: How do LLMs perform on the top 50 common and rare symptom datasets, and how does performance vary across these cases?
\end{enumerate}

\subsection{Experimental Setting}
\subsubsection{Dataset Description}
In the evaluation, we utilized the \ourdata\ dataset, comprising \ourdata-Full, \ourdata-Common-50, and \ourdata-Rare-50, as introduced in Section 3. These subsets enable comprehensive evaluation of LLM performance across both frequently and infrequently observed symptoms, providing insights into model robustness and adaptability. More details on the dataset and its creation can be found in Section 3.

% Based on the annotation process in Section 3, we categorized the annotated dataset, \ourdata-full, into two additional subsets: \ourdata-common-50 and \ourdata-rare-50. Symptoms were ranked by their frequency of occurrence within \ourdata-full, with the top 50 most frequently mentioned symptoms selected for \ourdata-common-50 and the 50 least frequently mentioned symptoms constituting \ourdata-rare-50. Each report in \ourdata-full was then evaluated for inclusion in these subsets, based on whether it contained at least one symptom from the respective lists. This categorization allows for a comprehensive evaluation of LLM performance across frequently and infrequently observed symptoms, offering deeper insights into the models' robustness and adaptability. \textcolor{red}{this section can be simplified, just briefly introduce. you can mention more details of the data can be found in section 3.}

\begin{table}[!tp]
    \centering
    \resizebox{1.0\linewidth}{!}{
    \begin{threeparttable}
    \caption{MATCH stage results assessing the quality and semantic accuracy of original symptom mentions extracted by different models. Bold values indicate the best performance for each prompting method, while italic values show the overall best performance across methods.}
    \label{tab:second_stage_eval}
    \begin{tabular}{ p{1.2cm}p{2.5cm}p{1.5cm}p{1.5cm}p{1.5cm}}
    \toprule
     \bf Prompt Type &\bf Models&\bf BLEU &\bf Fuzzy &\bf OpenAI Similarity \\
     \midrule
     & Jackalope-7b&0.238  & 0.728  & 0.604\\
     & Llama-2-13b-chat& 0.197 & 0.638  & 0.501\\
     TASI& gpt-3.5-turbo& 0.320  & 0.725  & 0.665\\
     & gpt-4-turbo& 0.377  & 0.770  & 0.721\\
     & gpt-4o& \textbf{0.382}  & \textbf{0.788}  & \textbf{0.725}\\
     \midrule
     & Jackalope-7b&0.182  & 0.621 & 0.501 \\
     & Llama-2-13b-chat& 0.214 & 0.717  & 0.541 \\
     TACO& gpt-3.5-turbo& 0.300  & 0.710  & 0.625\\
     & gpt-4-turbo& \textit{\textbf{0.465}}  & \textit{\textbf{0.862}}  & \textit{\textbf{0.775}} \\
     & gpt-4o& 0.435  & 0.856 & 0.766 \\
     
    \bottomrule
    \end{tabular}
    
    \end{threeparttable}
}
\end{table}

\subsubsection{Evaluation Metrics} 
% \textcolor{red}{describe all the evaluation metrics }
% need precision recall details for each stage and more details for each method.
% need reference for each method as well

% The evaluation metrics used to assess model performance are structured according to the two stages of the evaluation process. In the first stage, which focuses on the distillation and evaluation of linking, we employ Exact Match (EM) to measure accuracy by verifying if the extracted terms exactly match the human-annotated terms. Fuzzy Match accounts for minor discrepancies by allowing approximate matches between terms. The combined EM and Fuzzy Match approach is initially applied, where Exact Match is followed by Fuzzy Match for unmatched terms to ensure accurate matching. Precision and recall metrics quantify the accuracy and completeness of term linking.

% In the second stage, which evaluates the quality of extraction, we assess the accuracy of the original mentions generated by each model for the linked terms. This involves comparing these mentions to human labels and calculating similarity scores based on various evaluation metrics. BLEU Score measures the precision of n-grams between the generated and reference mentions, while Cosine Similarity evaluates the similarity between the generated and reference mentions using vector representations. The embedding for calculating cosine similarity is generated from OpenAI Embedding models.

The evaluation metrics used to assess model performance are structured according to the two stages of the evaluation process defined in Section \ref{eval-tasks}. Across both stages, we employ the following metrics: Exact Match (EM) \cite{em}, Fuzzy Match \cite{fuzzy}, Precision \cite{precisionrecall}, Recall \cite{precisionrecall}, BLEU Score \cite{bleu}, and Cosine Similarity \cite{cosine}. Each metric provides a unique perspective on the models' performance in extracting and linking adverse event symptoms and their original mentions.

In the first stage (LINK), the evaluation focuses on term linking, with Exact Match (EM), Fuzzy Match, Precision, and Recall serving as the primary metrics. EM measures accuracy by verifying if the extracted terms exactly match the human-annotated terms. Fuzzy Match accounts for minor discrepancies by allowing approximate matches between terms. The combined EM and Fuzzy Match approach is applied ultimately, where Exact Match is followed by Fuzzy Match for unmatched terms, ensuring a more comprehensive and accurate evaluation. Precision and recall metrics quantify the accuracy and completeness of term linking, providing insights into the models' ability to capture and associate relevant terms with their respective medical codes.

In the second stage (MATCH), the evaluation centers on the quality of the original mentions generated by the models. 
% Metrics employed in this stage include BLEU Score, Fuzzy Match, and Cosine Similarity. BLEU Score measures the precision of n-grams between the generated and reference mentions, while Fuzzy Match accounts for slight variations in wording that retain semantic similarity. Cosine Similarity evaluates the semantic coherence between the generated and reference mentions using vector representations. 
we assess mention quality using BLEU, Fuzzy Match, and Cosine Similarity-measuring n-gram precision, tolerating minor wording differences, and evaluating semantic coherence via vector embeddings, respectively.
The embeddings for cosine similarity are derived from OpenAI Embedding models, ensuring robust semantic comparisons.

% \noindent
\subsubsection{Implementation Details}
% \textcolor{red}{briefly describe the implementation details}

In our model implementation, we utilized the Jackalope-7B and Llama2-13B architectures, employing models sourced from The Bloke via Huggingface. Additionally, for GPT models, we accessed OpenAI APIs to obtain embeddings and inference results. To regulate the output length, we configured the parameter max\_new\_token to a value of 256, while adjusting the temperature within the range of 0.3 to 0.5 to optimize performance. Our experiments were conducted on hardware comprising two Nvidia RTX A5000 GPUs with 298.5GB disk space and 104GB RAM. The processing time for each model, spanning from obtaining inference results to generating evaluations, remained within 2 hours, ensuring timely and efficient execution.
% In our implementation, we employed Jackalope-7B and Llama2-13B models from The Bloke via Huggingface, alongside OpenAI APIs for GPT models. We controlled output length with max-newtoken set to 256 and adjusted temperature between 0.3 to 0.5 for optimal results. Experiments ran on two Nvidia RTX A5000 GPUs with 298.5GB disk space and 104GB RAM. Processing time for each model, from inference to evaluation, was under 2 hours.

\begin{figure*}[!tp]
	\centering
	\begin{subfigure}[b]{0.3\textwidth}
		\includegraphics[width=\textwidth]{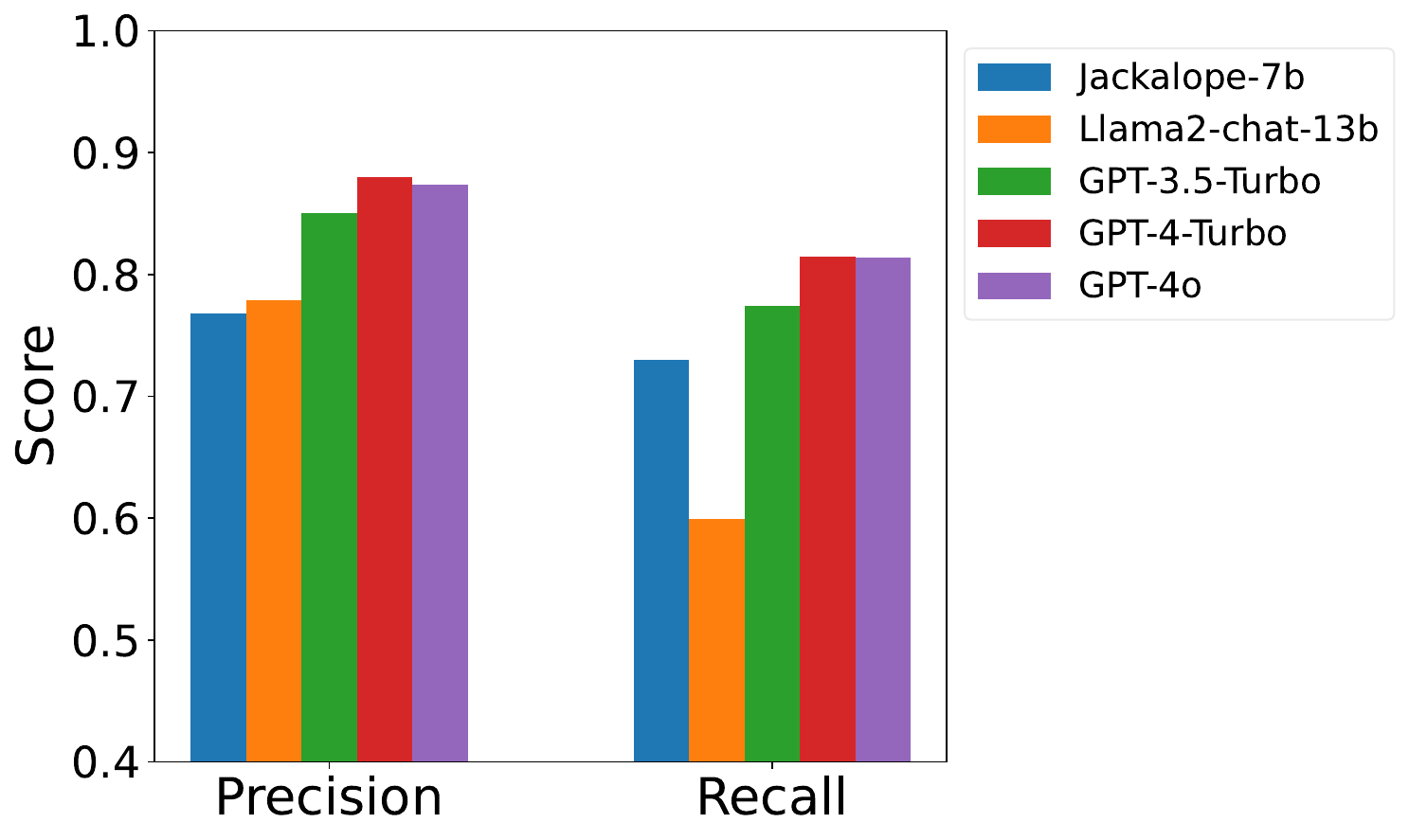}
		\vspace{-5mm}
		\caption{Top 50 TASI Prompting}
		\label{fig:common_rare_case_analysis_seq_a}
	\end{subfigure}
	\begin{subfigure}[b]{0.3\textwidth}
		\includegraphics[width=\textwidth]{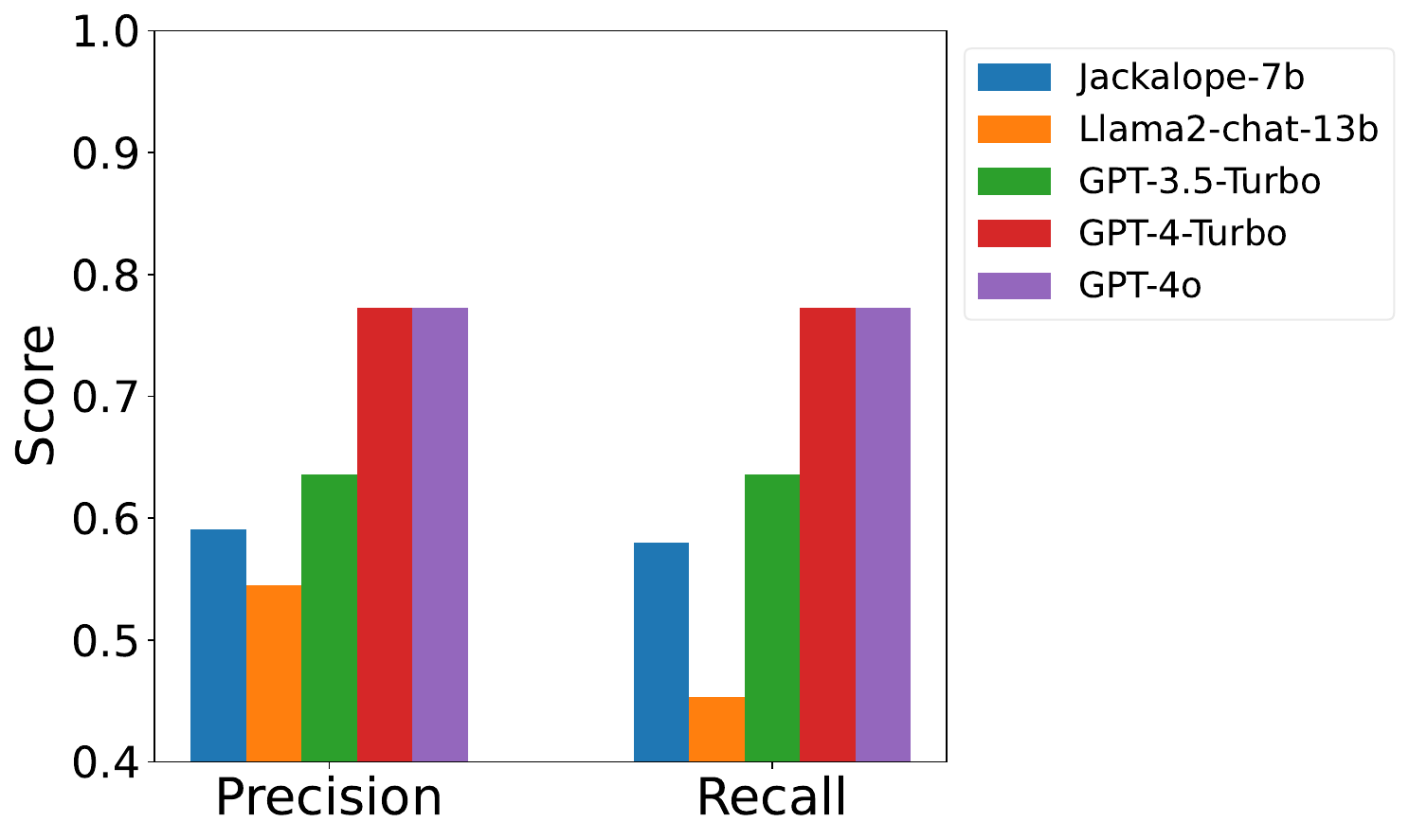}
		\vspace{-5mm}
		\caption{Bottom 50 TASI Prompting}
		\label{fig:common_rare_case_analysis_seq_b}
	\end{subfigure}
        \begin{subfigure}[b]{0.3\textwidth}
		\includegraphics[width=\textwidth]{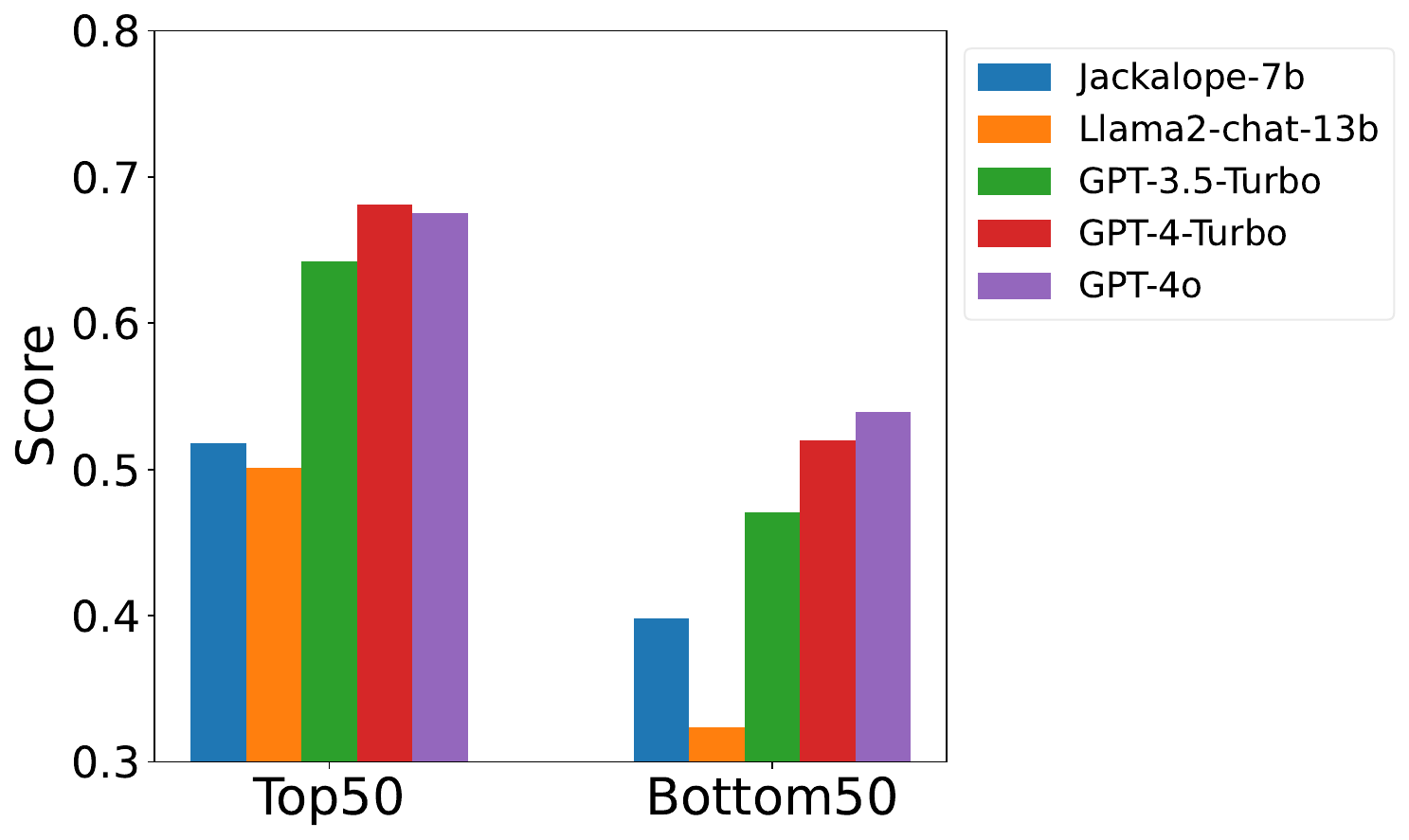}
		\vspace{-5mm}
		\caption{Similarity TASI Prompting}
		\label{fig:common_rare_case_analysis_seq_c}
	\end{subfigure}
	
	% \vspace{-4mm}
	\caption{Common Rare Case Analysis with TASI Prompting on \ourdata }
	% \vspace{-3mm}
	\label{fig:common_rare_case_analysis_seq}
\end{figure*}

\begin{figure*}[!tp]
	\centering
	\begin{subfigure}[b]{0.3\textwidth}
		\includegraphics[width=\textwidth]{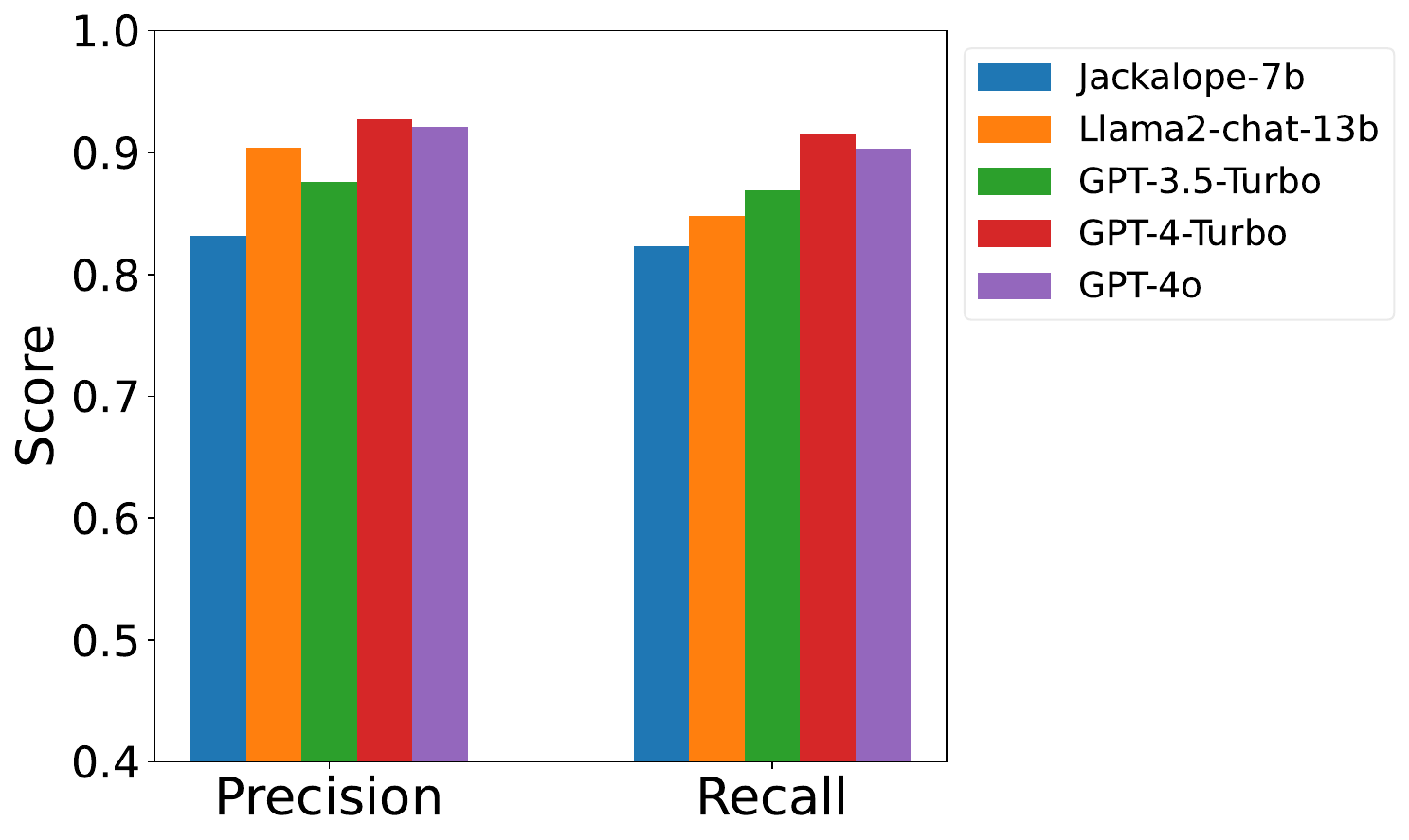}
		\vspace{-5mm}
		\caption{Top 50 TACO Prompting}
		\label{fig:common_rare_case_analysis_int_a}
	\end{subfigure}
	\begin{subfigure}[b]{0.3\textwidth}
		\includegraphics[width=\textwidth]{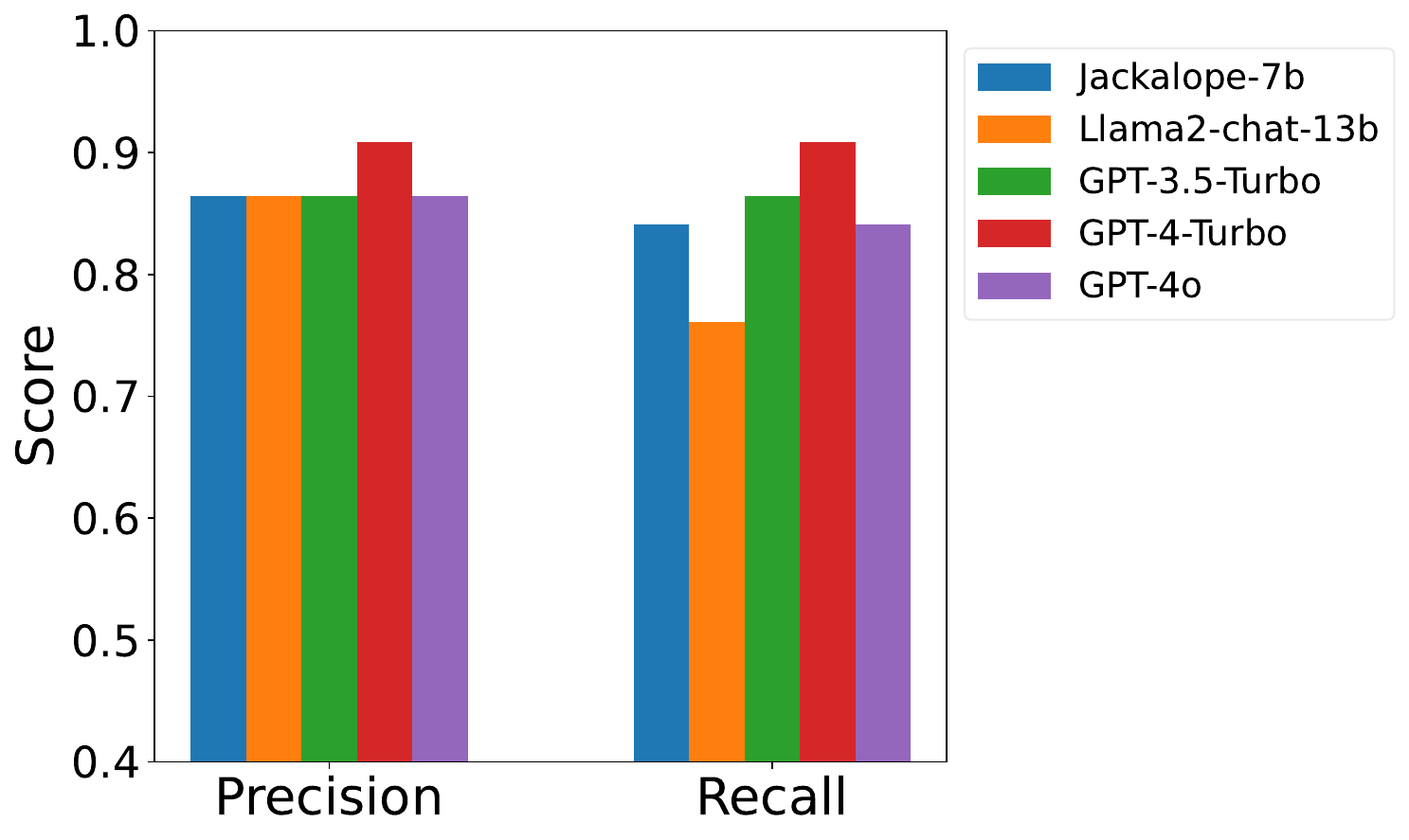}
		\vspace{-5mm}
		\caption{Bottom 50 TACO Prompting}
		\label{fig:common_rare_case_analysis_int_b}
	\end{subfigure}
	\begin{subfigure}[b]{0.3\textwidth}
		\includegraphics[width=\textwidth]{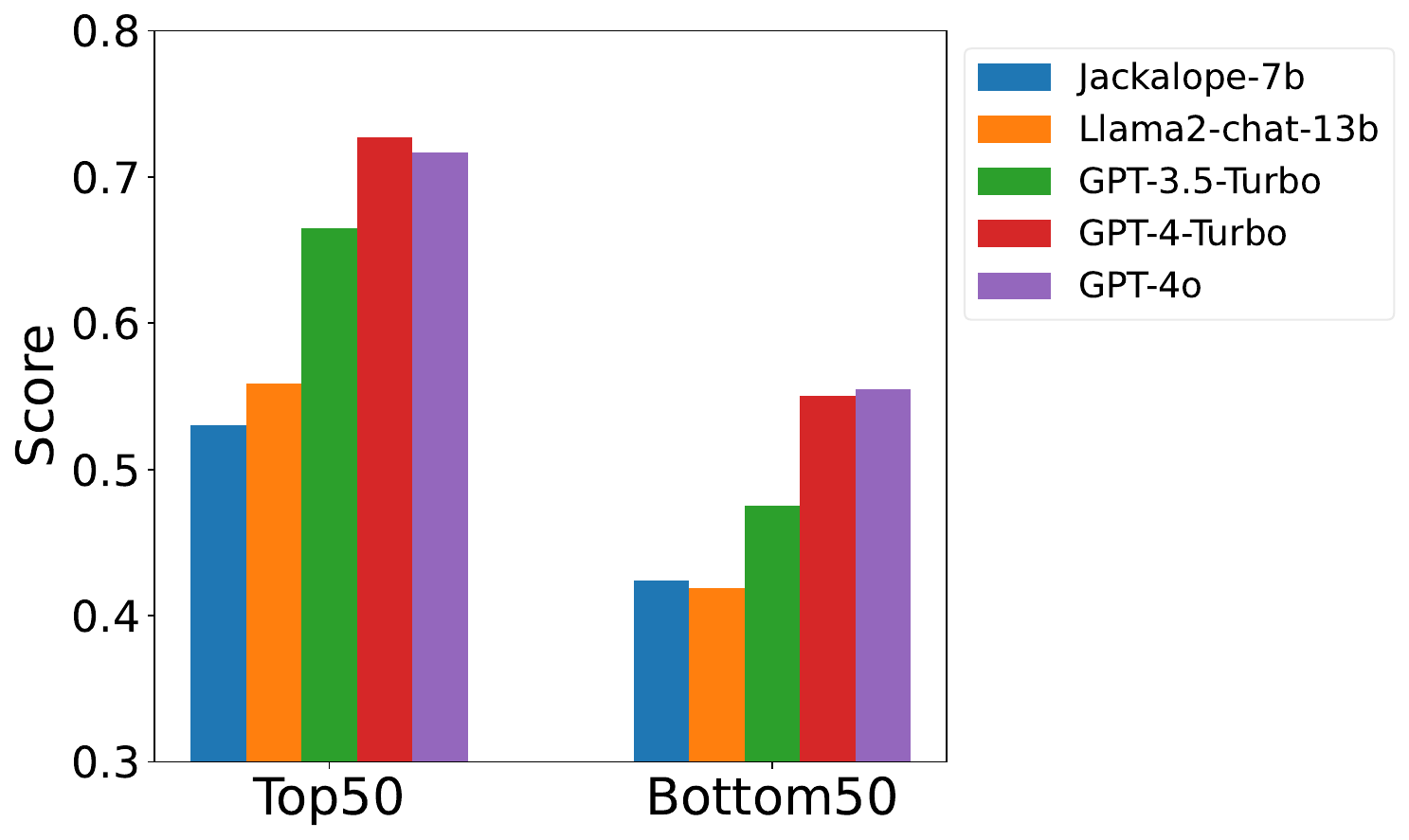}
		\vspace{-5mm}
		\caption{Similarity TACO Prompting}
		\label{fig:common_rare_case_analysis_int_c}
	\end{subfigure}
	
	% \vspace{-4mm}
	\caption{Common Rare Case Analysis with TACO Prompting on \ourdata }
	% \vspace{-3mm}
	\label{fig:common_rare_case_analysis_int}
\end{figure*}

\subsection{Experimental Results}

% \textcolor{red}{based on the results we have, have one subsection for each of them and analyze the results.}

\subsubsection{\textbf{Prompt Variation (RQ1)}}
To investigate the capacities of TACO prompting method, we compare TACO and TASI performances across both stages of evaluation. The results from Table \ref{tab:first_stage_eval} and Table \ref{tab:second_stage_eval} consistently demonstrate TACO's overall superiority in enhancing performance across most models in handling complex linking and extraction tasks.

In the \textit{LINK stage} shown in Table \ref{tab:first_stage_eval}, TACO achieves higher precision and recall scores for most models, with GPT-4o and GPT-4-Turbo consistently leading in performance. These results underline TACO’s ability to handle complex linking tasks effectively. However, Jackalope-7b exhibits a unique trend, where TASI slightly outperforms TACO for precision. This anomaly suggests that smaller models like Jackalope-7b may benefit from the sequential nature of TASI due to their limited capacity to handle the integrated context provided by TACO.
% TACO generally yields higher precision and recall, especially for GPT-4o and GPT-4-Turbo, while Jackalope-7b shows slightly better precision with TASI, indicating that smaller models may benefit from sequential prompting.

In the \textit{MATCH stage} displayed in Table \ref{tab:second_stage_eval}, TACO continues to outperform TASI across BLEU, fuzzy match, and similarity scores, particularly for advanced models like GPT-4o and GPT-4-Turbo. For instance, GPT-4-Turbo achieves a BLEU score of 0.465 and a fuzzy match score of 0.862, showcasing its ability to generate coherent and accurate mentions under TACO prompting method. However, models such as GPT-3.5-Turbo and Jackalope-7b exhibit a slight decline in BLEU and similarity scores with TACO. While the reasons for this decline are not entirely clear, it suggests potential limitations in how these models process the additional context provided by TACO.

Overall, TACO emerges as the more effective strategy, particularly for advanced LLMs like GPT-4o and GPT-4-Turbo, which leverages its integrated design to achieve robust and accurate performance. The results reaffirm TACO’s ability to streamline complex tasks and enhance both precision and recall, making it a superior prompting method for our task.

% These models consistently outperform others across various metrics, particularly in TACO prompt scenarios, which suggests that combining extraction and linking in a single prompt enhances model performance. 

\subsubsection{\textbf{Model Performance (RQ2)}}
To evaluate the impact of different LLMs on vaccine adverse symptom coding, we compared their performance across two evaluation stages: the LINK stage and the MATCH stage. As shown in Tables \ref{tab:first_stage_eval} and \ref{tab:second_stage_eval}, advanced LLMs such as GPT-4-Turbo and GPT-4o consistently outperform smaller models like Jackalope-7b and Llama2-13b, demonstrating their superior ability to handle the complexities of coding vaccine adverse symptoms.

In the \textit{LINK stage}, which evaluates how well models align symptoms with suggested terms, GPT-4-Turbo achieves near-perfect scores under TACO prompting, with an EM Precision of 0.999 and EM Recall of 0.998, reflecting its ability to both accurately and comprehensively code symptoms. Similarly, GPT-4o achieves exceptional performance under TASI prompting, with the highest EM-Fuzzy Precision (0.921) and EM-Fuzzy Recall (0.902). These results highlight the advanced contextual understanding of these models, enabling them to reliably handle the linking of symptoms to standardized terms.

In the \textit{MATCH stage}, which assesses the semantic accuracy and coherence of original mentions, GPT-4-Turbo again leads under TACO prompting with a BLEU score of 0.465 and a cosine similarity score of 0.775. These findings confirm that advanced models not only excel at linking symptoms to standardized terms but also maintain semantic accuracy and coherence in their outputs. In contrast, smaller models like Jackalope-7b and Llama2-13b demonstrate consistently lower performance, with Jackalope-7b showing significant difficulty in generating coherent mentions, particularly under TACO prompting.

\begin{table*}[!tp]
    \centering
    \caption{Original Mention vs Model Output from GPT4-turbo. This table compares human-annotated and model-extracted symptoms for six selected standard symptoms, with three examples each randomly chosen from the top 50 common and rare symptoms. The counts in parentheses indicate the frequency of mentions for each term. }
    % Common and Rare Symptoms: Human vs Model Mentions (GPT-4 Turbo)
    \begin{tabular}{cp{6.0cm}cp{7.5cm}}
        \toprule
        \bf Standard Symptoms &\bf  Human Annotated Symptoms &&\bf  Model Extracted Symptoms \\
        \midrule
       
        Fatigue & fatigue (59), tiredness (20), exhaustion (3) && fatigue (55), tiredness (23), exhaustion (6), weakness (3), wiped out (1)\\
        \cmidrule{2-4}
        Pyrexia & pyrexia (9), fever (102), elevated/inc temp (1) & &pyrexia (2), fever (108), elevated/inc temp (1), low grade temp (1), Temp of 103 degrees (3)\\
        \cmidrule{2-4}
        Dizziness & Dizziness (33), light headed (5), woozy (2), vertigo (2) && Dizziness (33), light headed (5), woozy (2), wobbly legs (1)\\
         \midrule
         Eye Irritation & eye irritation (1), burning eyes (1) & &burning eyes (1)\\
         \cmidrule{2-4}
        Facial Spasm & facial muscle spasm (1) & &facial muscle spasm (1) \\
        \cmidrule{2-4}
        Blister & blisters (1), blister (1) && blisters (1)\\
       
        \bottomrule
    \end{tabular}
    
    \label{tab:term expression}
\end{table*}

An interesting observation is the relative performance of Jackalope-7b, which surpasses Llama2-13b in the LINK stage under TASI. This unexpected trend could stem from Jackalope-7b’s simpler architecture and focused training, which might align better with the structured nature of TASI prompts. However, this advantage diminishes in the TACO scenario, where the added contextual complexity benefits more advanced models like GPT-4-Turbo and GPT-4o. This suggests that larger and more advanced LLMs are better equipped to handle the challenges of contextual integration inherent in TACO prompting.

Overall, the results demonstrate that GPT-4-Turbo and GPT-4o are the most effective models for vaccine adverse symptom coding. Their advanced architectures and contextual capabilities enable them to achieve consistently high performance across both evaluation stages, making them the most reliable choices for addressing the complexities of this task.

\subsubsection{\textbf{Common Rare Symptoms Analysis (RQ3)}} 
To evaluate the performance of LLMs on datasets containing the top 50 most common and bottom 50 least common symptoms, we conducted analyses using both TASI and TACO prompting methods. Figures \ref{fig:common_rare_case_analysis_seq} and \ref{fig:common_rare_case_analysis_int} provide insights into how models handle frequent versus infrequent adverse symptoms, with a focus on model performance under each prompting strategy before comparing the methods.

Under \textit{TACO prompting}, as shown in Figure \ref{fig:common_rare_case_analysis_int}, models like GPT-4o and GPT-4-Turbo consistently exhibit higher precision and recall for the top 50 most common symptoms (Figure 4a). This demonstrates their ability to accurately and comprehensively capture frequent symptoms due to their advanced architectures and extensive training. However, for the bottom 50 least common symptoms (Figure 4b), performance varies more significantly. GPT-4-Turbo maintains relatively robust recall and precision, while GPT-4o and Llama2-13b-chat experience a noticeable decline in both metrics. This decrease suggests that rare symptoms, despite their distinctiveness, are challenging to extract consistently under TACO prompting, likely due to their sparse representation in training data. The similarity scores (Figure 4c) further support these observations, with GPT-4o and GPT-4-Turbo achieving the highest semantic accuracy for both common and rare symptoms, while smaller models like Llama2-13b-chat perform less effectively.

Under \textit{TASI prompting}, as depicted in Figure \ref{fig:common_rare_case_analysis_seq}, similar trends emerge, but the recall values for rare symptoms exhibit a sharper decline compared to TACO prompting (Figure 5b). For the top 50 most common symptoms (Figure 5a), all models maintain high precision, with GPT-4o and GPT-4-Turbo again leading in recall. However, for the bottom 50 rare symptoms, recall values decrease more dramatically for models like Llama2-13b-chat and GPT-4o, indicating that TASI’s sequential structure is less effective at capturing rare terms. Precision scores for rare symptoms also decline under TASI, although not as significantly as recall. The similarity scores (Figure 5c) show consistent results, with GPT-4o and GPT-4-Turbo outperforming smaller models, particularly in the rare case analysis.

When comparing the two prompting methods, TACO consistently demonstrates advantages over TASI for capturing common symptoms, with higher recall and precision for most models. For rare symptoms, TACO prompting provides better consistency in recall and precision across advanced models like GPT-4-Turbo and GPT-4o, although challenges remain for smaller models. The integrated contextual design of TACO prompting likely allows models to handle both frequent and infrequent patterns more effectively, while TASI’s sequential structure introduces limitations, particularly for rare cases.

Overall, the comparison reveals that advanced models such as GPT-4o and GPT-4-Turbo consistently outperform smaller models under both prompting methods. However, TACO prompting offers clear advantages in handling the variability of both common and rare adverse symptoms, highlighting its effectiveness as a strategy for vaccine adverse symptom coding tasks.

\begin{table*}[!tp]
\centering
\vspace{-3mm}
\caption{Illustrative Example of Model Prediction Comparison. This table compares the outputs of various models against the gold labels for a selected VAERS report, illustrating the models' abilities to identify and link symptoms from the given text description to standardized terms. Text in the same color refers to the same symptom, while red text indicates incorrect predictions compared to the gold labels.}

  \resizebox{\linewidth}{!}{
 
  \begin{tabular}{lp{0.9\linewidth}}
    \toprule
    \bf Text Descriptions & 24 hours after vaccines had temp of 103.2 degrees. Gave Tylenol. On Tues., lack of appetite. On Wed, blotchy rash on face \& body. On Thurs, rash lessened. On Fri, no rash. Fever lasted about 24 hours. Has had cold for approx. 3 weeks/runny nose, occasional fever). Directed to inform physician of vaccines and events. \\
    \cmidrule(lr){2-2}
    \bf Gold Labels & \textcolor{orange}{Pyrexia: [Fever]}, \textcolor{teal}{Decreased appetite: [lack, of, appetite]}, \textcolor{violet}{Rhinorrhoea: [weeks/runny, nose]}, \textcolor{cyan}{Rash: [blotchy, rash, on, face, \& body]}, \textcolor{brown}{Rash macular: [blotchy, rash]} \\

    \toprule
    \bf Model & \bf Output \\
    \midrule
    \bf Jackalope-7b & \textcolor{orange}{Pyrexia: [fever, temp of 103.2 degrees, \textcolor{red}{lack of appetite, on thursday, no rash, on friday}]}, \textcolor{teal}{Decreased appetite: [\textcolor{red}{on tuesday, on wednesday, on friday}]}, \textcolor{violet}{Rhinorrhoea: [cold for approx. 3 weeks/runny nose]}, \textcolor{cyan}{Rash: [blotchy rash on face \& body, \textcolor{red}{on wednesday, on thursday, on friday}]}, \textcolor{brown}{Rash macular: [\textcolor{red}{rash on face \& body, rash on wednesday, rash on thursday, rash on friday}]} \\
    \cmidrule(lr){2-2}
    \bf Llama2-chat-13b & \textcolor{orange}{Pyrexia: [fever]}, \textcolor{teal}{Decreased appetite: [lack of appetite]}, \textcolor{violet}{Rhinorrhoea: [\textcolor{red}{none}]}, \textcolor{cyan}{Rash: [blotchy rash, rash lessened, \textcolor{red}{no rash}]}, \textcolor{brown}{Rash macular: [\textcolor{red}{none}]} \\
    \cmidrule(lr){2-2}
    \bf GPT-3.5-Turbo & \textcolor{orange}{Pyrexia: [\textcolor{red}{temp}, fever]}, \textcolor{teal}{Decreased appetite: [lack of appetite]}, \textcolor{violet}{Rhinorrhoea: [\textcolor{red}{cold}, runny nose]}, \textcolor{cyan}{Rash: [blotchy rash]}, \textcolor{brown}{Rash macular: [\textcolor{red}{none}]} \\
    \cmidrule(lr){2-2}
    \bf GPT-4-Turbo & \textcolor{orange}{Pyrexia: [temp of 103.2 degrees, Fever lasted about 24 hours]}, \textcolor{teal}{Decreased appetite: [lack of appetite]}, \textcolor{violet}{Rhinorrhoea: [runny nose]}, \textcolor{cyan}{Rash: [blotchy rash on face \& body]}, \textcolor{brown}{Rash macular: [\textcolor{red}{none}]} \\
    \cmidrule(lr){2-2}
    \bf GPT-4o & \textcolor{orange}{Pyrexia: [temp of 103.2 degrees, fever lasted about 24 hours, occasional fever]}, \textcolor{teal}{Decreased appetite: [lack of appetite]}, \textcolor{violet}{Rhinorrhoea: [runny nose]},\textcolor{cyan}{Rash: [blotchy rash on face \& body, rash lessened, \textcolor{red}{no rash}]}, \textcolor{brown}{Rash macular: [\textcolor{red}{none}]} \\
    \bottomrule
  \end{tabular}
  }
  
  % \vspace{-3mm} 
  \label{tab:model_comparison}
\end{table*}

\subsubsection{\textbf{Case Study: Common vs Rare Mention}}
Table \ref{tab:term expression} evaluates the ability of GPT-4 Turbo to capture common and rare terms in clinical text. The selected terms—"Eye Irritation," "Facial Spasm," "Blister," "Fatigue," "Pyrexia," and "Dizziness"—are derived from the top 50 (common) and bottom 50 (rare) terms in the dataset. This selection provides a balanced analysis of the model's performance across varying levels of term frequency.

For common terms like "Fatigue" and "Dizziness," GPT-4 Turbo enriched its outputs with additional contextual details, such as "wiped out" for "Fatigue" and "wobbly legs" for "Dizziness." While this demonstrates the model's ability to infer related terms, the overgeneralization occasionally introduced extraneous phrases that deviated from the original clinical mention. In the case of "Pyrexia," the model generated descriptors such as "low-grade temp" and "temp of 103 degrees," which, while adding specificity, risked fragmenting data due to overly detailed outputs.

Rare terms, on the other hand, posed greater challenges. For example, "Eye Irritation" was simplified to "burning eyes," reflecting a preference for informal phrasing over formal clinical terminology. Similarly, terms like "Facial Spasm" and "Blister" were extracted correctly but lacked variation, suggesting limited diversity in the model's outputs for rare terms. 
% A potential explanation for these omissions could be the sparsity of rare case reports in the training data. The underrepresentation of such terms may lead the model to favor informal phrases due to greater familiarity, even when both formal and informal phrases appear in the same report. This tendency suggests that the model might rely on heuristic patterns developed during training, prioritizing commonly observed terms over rare or specific phrasing.
These issues may stem from the sparse representation of rare cases in the training data, causing the model to favor familiar informal phrases and rely on heuristics that prioritize common terms over rare, specific ones.

% In summary, GPT-4 Turbo demonstrates robust performance in recognizing and elaborating on common terms but faces challenges in preserving formal phrasing for rare terms. The model's reliance on sparse data and potential preference for familiar terms highlights the importance of refining prompts and incorporating post-processing techniques to improve handling of underrepresented terms in clinical text processing.
In summary, GPT-4 Turbo performs well with common terms but struggles to preserve formal phrasing for rare ones. This reliance on sparse data underscores the need to refine prompts and add post-processing to better handle underrepresented terms.

\subsubsection{\textbf{Case Study: Mention Discrepancy}}
Table \ref{tab:model_comparison} compares outputs from five models(Jackalope-7b, Llama2-chat-13b, GPT-3.5-Turbo, GPT-4 Turbo, and GPT-4o) with gold-standard annotations, revealing significant differences in capturing nuanced and rare mentions.

Advanced models like GPT-4 Turbo and GPT-4o consistently provided detailed and contextually enriched outputs. For instance, GPT-4o added specific details such as "occasional fever," closely aligning with the gold standard while offering richer context. Jackalope-7b, on the other hand, tended to over-annotate by introducing temporal phrases like "on Wednesday" and "on Friday," which, while detailed, often introduced unnecessary noise. Llama2-chat-13b struggled significantly, failing to identify certain mentions, particularly less frequent ones like "Rash Macular," highlighting its limitations in capturing rare or complex terms.

% For common terms such as "Decreased Appetite," most models, except Jackalope-7b, aligned closely with the gold standard, demonstrating their ability to handle frequent terms effectively. However, the models showed varying degrees of success in capturing rare terms like "Rash Macular." While Jackalope-7b provided multiple detailed mentions, other models such as Llama2-chat-13b and GPT-3.5-Turbo marked the term as "none," underscoring the challenges of identifying rare terms across different architectures.
For common terms like 'Decreased Appetite,' all models except Jackalope-7b closely matched the gold standard. However, performance on rare terms like 'Rash Macular' varied: Jackalope-7b offered multiple details, while Llama2-chat-13b and GPT-3.5-Turbo often returned 'none.'

This analysis reinforces a clear trend: higher-capacity models generally excel in capturing nuanced terms, whereas smaller models struggle with rare and complex ones. Even the most advanced models miss infrequent terms occasionally, highlighting an ongoing need for refined prompting strategies and post-processing pipelines to ensure clinical relevance and consistency. 

\section{Conclusions}

In this paper, we addressed the challenges of accurate symptom coding from unstructured clinical data, particularly focusing on the nuanced extraction and linking of symptoms in vaccine safety reports such as those in VAERS. To benchmark the performance of LLMs on this tailored task, we introduced \ourdata, a human-annotated dataset that captures both common and rare symptoms, providing a robust foundation for evaluating adverse event extraction and linking tasks. Our proposed TACO prompting framework unifies symptom extraction and linking into a single process, significantly improving contextual accuracy and reducing information loss typically associated with traditional separate workflows. In addition, we introduced a two-stage evaluation framework, comprising the LINK and MATCH phases, which enables a detailed assessment of model performance by focusing on both symptom linking and original mention matching. This framework revealed key differences in how models handle common versus rare symptoms, offering a nuanced understanding of LLM capabilities. Through extensive experiments with several state-of-the-art LLMs, 
% including Llama2-chat, Jackalope-7b, GPT-3.5 Turbo, GPT-4 Turbo, and GPT-4o, 
we demonstrated the superior performance of TACO, emphasizing the impact of task-specific prompt design on model accuracy. Our findings not only highlight the effectiveness of TACO prompting in enhancing model performance but also underscore its potential as a flexible framework for developing tailored coding tasks in clinical natural language processing. Future work will explore broader applications of TACO, aiming to develop more innovative and impactful solutions in the medical field.

% \newpage

\begin{acks}
This work was supported in part by the US National Science Foundation grant IIS-2245907, 2047843, 2437621, and the startup funding from the Stevens Institute of Technology.
\end{acks}

\balance
\bibliographystyle{ACM-Reference-Format}
\bibliography{ref}

% \let\clearpage\relax
\end{document}